\documentclass[a4paper]{article}
\usepackage{arxiv}
\usepackage{graphicx}
\usepackage{subfigure,comment}
\usepackage{booktabs}
\usepackage[colorlinks=false, pdfborder={0 0 0}]{hyperref}
\usepackage{algorithm, algorithmic}
\usepackage{natbib}
\usepackage{url}
\usepackage{hyperref} 
\usepackage{amsmath}
\usepackage{amssymb}
\usepackage{mathtools}
\usepackage{amsthm}
\usepackage{subcaption}
\usepackage{cleveref}

\usepackage{setspace}
\usepackage{color}
\usepackage{enumitem}

\usepackage{mathptmx}
\usepackage{multicol}
\usepackage{lipsum}
\usepackage[normalem]{ulem}
\useunder{\uline}{\ul}{}
\usepackage{multirow}

\theoremstyle{plain}

\theoremstyle{definition}

\theoremstyle{remark}

\usepackage{xspace}

\newcommand\gaa{GENOME}
\newcommand\ga{GENOME\xspace}
\newcommand\gsa{GENOME+\xspace}
\newcommand\gsaa{GENOME+}
\newcommand\fullga{\textbf{GEN}etic \textbf{O}ptimization for \textbf{M}odel \textbf{E}volution\xspace}
\newcommand\github{\url{https://github.com/ZhangYiqun018/GENOME}}

\newcommand\huggingface{\url{https://huggingface.co/Estwld/GENOME-gemma-2b-it}}

\usepackage[textsize=tiny]{todonotes}
\usepackage{comment}

\title{{Nature-Inspired Population-Based Evolution of Large Language Models}}
\runningtitle{Nature-Inspired Population-Based Evolution of Large Language Models}

\author{
  \textbf{Yiqun~Zhang}$^{1~2~\ast}$ \quad 
  \textbf{Peng~Ye}$^2$  \quad
  \textbf{Xiaocui~Yang}$^1$ \quad
  \textbf{Shi~Feng}$^{1~\dag}$ \\
  \textbf{Shufei~Zhang}$^2$ \quad
  \textbf{Lei~Bai}$^2$ \quad
  \textbf{Wanli~Ouyang}$^2$ \quad 
  \textbf{Shuyue Hu}$^{2~\dag}$\\
  $^1$ \text{Northeastern University, China} \\
  $^2$ \text{Shanghai Artificial Intelligence Laboratory} \\
  \texttt{yiqunzhang@stumail.neu.edu.cn} \quad 
  \texttt{\{yangxiaocui,fengshi\}@cse.neu.edu.cn} \\ 
  \texttt{\{yepeng,bailei,ouyangwanli,hushuyue\}@pjlab.org.cn}
}
\date{   }
\begin{document}








\maketitle


\begin{abstract}
Evolution, the engine behind the survival and growth of life on Earth, operates through the population-based process of reproduction. 
Inspired by this principle, 
this paper formally defines a newly emerging problem---the population-based evolution of large language models (LLMs)---and introduces a novel framework. Starting with a population of parent LLMs, our framework enables the population to evolve through four key operations:  
(i) crossover, merging the weights of different parents to create offspring LLMs,
(ii) mutation, introducing small, random changes to model weights to foster diversity,
(iii) selection, prioritizing high-performing models,
and (iv) succession, transferring the learned experience from parent to offspring LLMs.
With only 200 samples per new task, the LLM population evolves rapidly to adapt to the task at hand, without any gradients. 
Experiments on 12 datasets show that our framework consistently outperforms existing multi-LLM merging and adaptation methods, achieving accuracy gains of up to $54.8\%$ over the best LLM in the initial population. 
Moreover, our framework allows for the evolution of LLMs across multiple new tasks simultaneously, scaling effectively with populations of up to 40 LLMs, and even zero-shot generalization to unseen held-out tasks.
We have open-sourced the code on GitHub$^\spadesuit$ and released the weights of 10 parent LLMs, fine-tuned from \textit{gemma-2-2b-it}, on HuggingFace$^\clubsuit$, enabling reproduction of our proposed framework using just a single 4090 GPU with 24GB memory, without any performance degradation.
\end{abstract}

\def\thefootnote{\dag}\footnotetext{Corresponding author.}
\def\thefootnote{$\spadesuit$}\footnotetext{\github}
\def\thefootnote{$\clubsuit$}\footnotetext{\huggingface}
\def\thefootnote{$\ast$}\footnotetext{Work done during the author’s internship at Shanghai Artificial Intelligence Laboratory.}

\section{Introduction}
For billions of years, nature has taught us a profound lesson: \emph{evolution} drives the survival and flourishing of all life on Earth, from the simplest single-celled organisms to complex ecosystems and human civilizations.
Evolution works not on individuals in isolation but on entire populations, where \emph{``survival of the fittest''} shape the story of life.
Now, imagine channeling these very principles not to biological species but to cutting-edge AI technology---what if we could harness population-based evolution to advance large language models (LLMs)?

There are multiple reasons for the population-based evolution of LLMs to be compelling, despite being under-explored.
First of all, unlike the relatively few pre-trained LLMs, there is a growing proliferation of fine-tuned ``expert'' models that are specialized to specific tasks\footnote{There are more than 170,000 expert models, trained through parameter-efficient fine-tuning (PEFT) techniques, available on Hugging Face~\citep{muqeeth2024learning}.}, laying a solid foundation for the concept of a population of LLMs.
Second, recent studies have shown that merging the weights of LLMs with different expertise can produce versatile models~\citep{yu2024dareties,goddard2024arcee,mavromatis2024packllmsmodelfusion}, sometimes even resulting in the emergence of new capabilities~\citep{yadav2024matters}. 
Most importantly, enabling the evolution of an LLM population could allow these models to dynamically adapt to new, unseen tasks that have not been explicitly trained on~\citep{huang2024lorahubefficientcrosstaskgeneralization,mavromatis2024packllmsmodelfusion}. This effectively recycles the substantial efforts and computing resources invested in training the original models. 
Moreover, it can open the door to democratizing foundation model development~\citep{akiba2024evolutionary}, encouraging a broader community to jointly advance the field of LLMs.

To this end, we introduce the formulation of population-based evolution of LLMs, and propose a novel framework that draws inspiration from the very basic units of life---genes. 
We conceptualize the weights of an LLM as analogous to genes. After all, just as genes biologically define who we are, the weights of LLMs define their capabilities and behaviors. 
When considering evolution in relation to genes, the seminal genetic algorithm (GA)~\citep{holland1992genetic} naturally comes to mind. 
We observe a fundamental connection between the population-based evolution of LLMs and GA---while the former seeks to adapt to a specific task through optimizing the LLMs' weights, the latter is a well-established approach for optimization problems.
Building on this connection, we reintroduce four key principles of the GA----fitness, crossover, mutation, and selection---within the context of population-based evolution of LLMs.
Specifically, \textbf{fitness} assesses how well a LLM performs (or fits) on a specific task,
\textbf{crossover} merges the weights of different LLMs, generating offspring models and enabling them to inherit advantageous features from parent models, 
\textbf{mutation} adds small random perturbations to the model weights, enabling the model to acquire new abilities, 
and \textbf{selection} selects high-performing models to survive, effectively driving the LLM population to evolve in a better direction.
We refer to our framework that embodies these four concepts, closely aligning with GAs, as \textbf{\ga} (\fullga).

\begin{figure*}
\centering
\includegraphics[width=\linewidth]{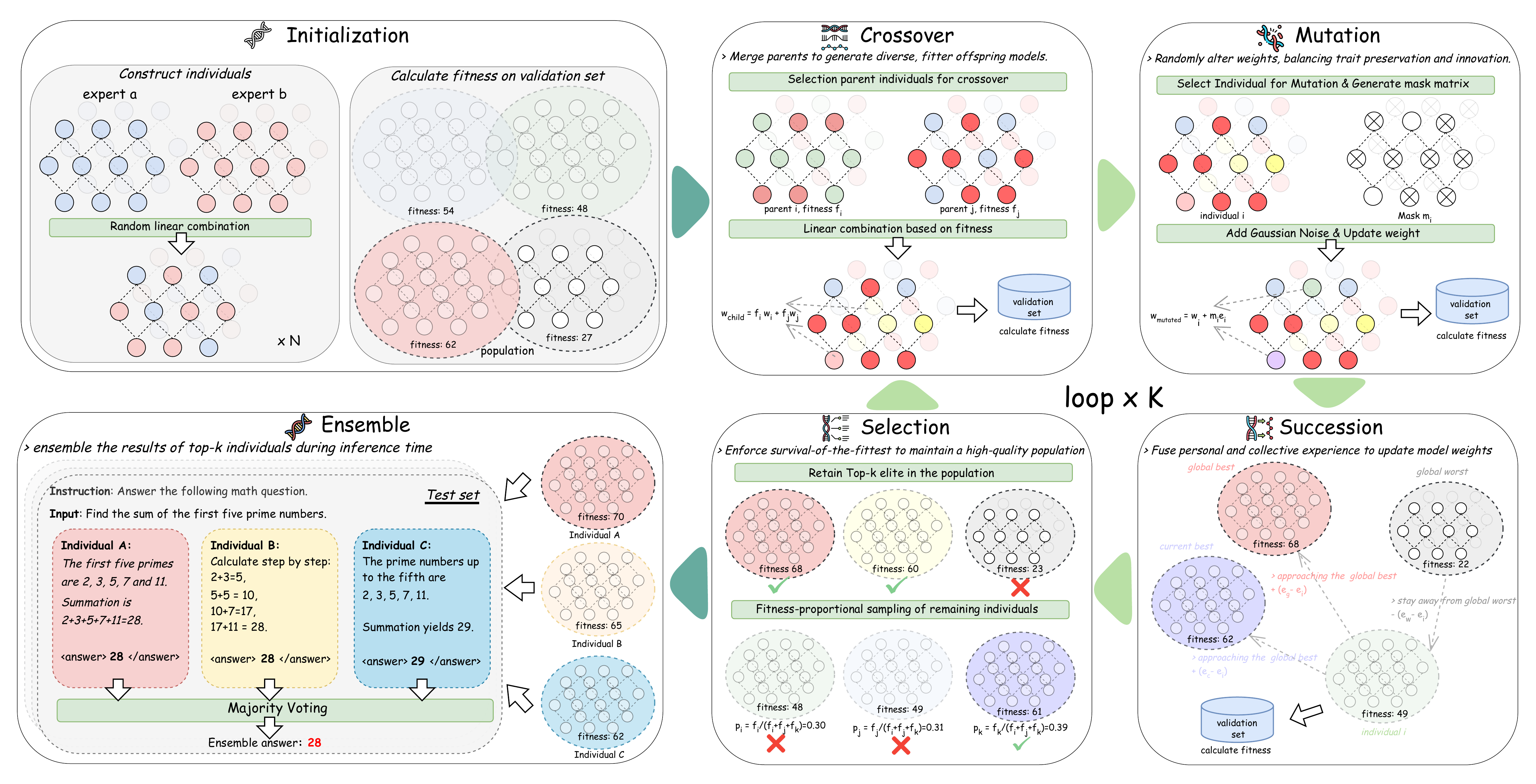}
\caption{GENOME+: a population-based evolutionary framework,   including \textbf{crossover}, \textbf{mutation}, \textbf{succession}, \textbf{selection} and \textbf{ensemble} operations, for LLMs.}
\label{fig: overview}
\end{figure*}

While connecting GA to LLMs is conceptually appealing, we acknowledge that traditional GA heavily emphasizes on reproduction. However, survival and progress in nature are not solely driven by reproduction. Humans and other species thrive not only by passing down genetic traits but also by learning from peers and previous generations, as well as through collective decision-making. 
Thus, to fully harness nature's evolutionary principles, we introduce two operations:  \textbf{succession}, which allows LLMs to learn from the best-performing LLM in the population while also drawing bitter lessons from the worst ones, and  \textbf{ensemble}, which  integrates the outputs of top-$k$ best performing models into a final answer.
As shown in Figure~\ref{fig: overview}, we extend \ga with succession and ensemble operations to create \textbf{\gsa}.
Compared to \ga, \gsa makes better use of the diversity inherent in a LLM population, mirroring the wisdom of the crowd. 
Despite their differences, both methods (collectively referred to as \textbf{\gaa(+)}) are easy to implement and enable gradient-free weight updates, eliminating the need for the traditionally time- and cost-intensive backpropagation process. This not only enhances accessibility but also significantly reduces the computational overhead of population-based evolution for LLMs.

We conduct extensive evaluations on 12 datasets spanning 7 categories, demonstrating that \gaa(+) consistently outperforms 7 existing methods, with only 200 samples. 
Notably, compared to the state-of-the-art Model Swarms~\citep{feng2024model}, \gsa achieves an average gain of 10.75\% and excels in reasoning-intensive tasks, improving by 36.3\% in DROP~\citep{dua2019dropreadingcomprehensionbenchmark} and 19.3\% in MGSM~\citep{shi2022languagemodelsmultilingualchainofthought}.
When evolving to simultaneously adapt to two tasks, \gaa(+) maintain stable improvement, while existing methods typically degrade. 
Furthermore, \gaa(+) are able to effectively generalize to unseen tasks without any sample, attaining an 11.79\% boost in cross-dataset transfer. 
Even as the population scales up to 40 LLMs, both methods continue to demonstrate performance improvements.
To assess computational efficiency, we reproduce key experiments on a single 4090 GPU with 24GB of memory, demonstrating that our method does not require extensive computational resources while maintaining consistent performance.
Last but not least, an ablation study confirms the positive contribution of each evolutionary operation. These results underscore the efficacy and adaptability of our population-based evolution framework, across diverse and challenging scenarios.

To summarize, our key contributions are as follows:
\vspace{-1.0em}
\begin{itemize}[leftmargin=*]
\setlength\itemsep{0em} 
    \item We formally define the LLM population-based evolution problem and present a novel conceptual framework that connects LLM weights to genes, task performance to individual fitness, and weight updates to evolutionary operations.
    
    \item We propose \ga and \gsa, which incorporates the key principles of traditional genetic algorithms as well as two operations: succession and ensemble. This showcases how traditional evolutionary algorithms can shed light on the population-base evolution of LLMs.
    
    \item We conduct extensive experiments in single task, multi-task domain, and zero-shot generalization settings, demonstrating \gsa’s advantages, verifying its scalability to 40-model populations, and confirming each operation’s necessity through ablation studies. Experiments show that our results can be reproduced on a 4090 GPU with 24GB of memory.
\end{itemize}

Overall, our work channels nature’s timeless lesson---population-based evolution---to the development of LLMs.
Embracing those evolutionary principles, we offer an easy-to-implement, gradient-free, and cost-efficient framework that enables a population of LLMs to quickly adapt to newly unseen tasks with minimal samples.
We hope our work can not only encourage exploration into the population-based evolution of LLMs, but also ignite imagination about what we can learn from the greatest teacher of all---Nature---as we push the boundaries of AI innovation to new heights.

\section{Methodology}
\subsection{LLM Population-based Evolution}
We define the \textbf{LLM population-based evolution} as an iterative optimization problem over a population of large language models (LLMs). Given a population:

\begin{equation}
P^{(t)} = \{x_i^{(t)}\}_{i=1}^{N},\quad x_i^{(t)} \leftrightarrow \mathbf{w}_i^{(t)} \in \mathbb{R}^{d}
\end{equation}

where each model \( x_i^{(t)} \) is represented by learnable parameters \( \mathbf{w}_i^{(t)} \), which may encompass either the full LLM weights or a subset thereof (such as LoRA or adapter parameters). The performance of each individual model on a given task \(T\) is quantified by a fitness function \(f(\mathbf{w}_i^{(t)}; T)\).

The goal is to evolve the population \(P^{(t)}\) over iterations \(t=0,1,\dots,K\) to optimize the aggregated fitness of the top-\(k\) individuals in the final population \(P^{(K)}\):
\begin{equation}
\max_{\{P^{(t)}\}_{t=0}^{K}}\; \frac{1}{k}\sum_{i=1}^{k} f(\mathbf{w}_{(i)}^{(K)};T),
\end{equation}
where \(\mathbf{w}_{(i)}^{(K)}\) denotes the parameters of the \(i\)-th best-performing individual in \(P^{(K)}\) ranked by fitness. The integer \(k\in[1,N]\) controls whether the optimization focuses solely on the best individual (\(k=1\)), or collectively improves a larger subset or the entire population (\(k>1\)).

\subsection{GENOME}
In nature, the evolution of biological populations follows intricate laws: genetic material creates diversity through \textbf{crossover} and \textbf{mutation}, while natural \textbf{selection} acts as an invisible sieve, retaining traits with stronger adaptability across generations. This ``crossover-mutation-selection" iterative mechanism allows species to continuously optimize their survival strategies in complex and changing environments. When we examine the performance of LLMs in specific tasks from the perspective of evolutionary principles, similar patterns emerge: each model is like an individual with unique traits, its weight parameters akin to a digital genetic sequence encoding the core ability to solve tasks, and its performance on specific tasks acts like the sieve of natural selection, determining its opportunities for ``survival" and ``reproduction".

Drawing on insights from biological evolution, we propose \ga (\fullga), which combines the evolutionary mechanisms of GA with the population dynamics of LLMs.
Here, the target task $T$ serves as the ``environment", and the \textbf{fitness function} $f$ is the performance on the validation set of $T$. 
This method begins with a population of $n$ homologous LLM expert models and creates an initial population $P^{(0)}=\{x_i\}^N_{i=1}$ via an \textbf{initialization} operation, 
where each individual's genes are represented by its \textbf{LoRA} weights \( \mathbf{w}_i \) for parameter-efficient fine-tuning. 
The \textbf{crossover} operation is employed to merge beneficial genes, while the \textbf{mutation} operation is utilized to increase population diversity. 
The \textbf{selection} operation simulates natural selection pressure by retaining advantageous individuals and facilitating the evolution of the model population. Algorithm \ref{alg: GA} outlines the evolution process of the population, which includes initialization, crossover, mutation, and selection, executed over $K$ iterations.

\paragraph{Initialization}
Starting with \(n\) LLM expert models \(\{x_i\}^n_{i}\), we construct a diverse initial population through random linear combinations of expert model weights. For each new individual $i$, we combine the LoRA parameters \(\mathbf{w}_a\) and \(\mathbf{w}_b\) from two expert models:
\begin{equation}
\mathbf{w}_i \leftarrow t \cdot \mathbf{w}_a + (1-t) \cdot \mathbf{w}_b, \quad t \sim U(0,1) 
\end{equation}
This process is repeated \(N\) times to form a population of size \(N\). 
We then evaluate each individual on a validation set using the fitness function to obtain fitness scores \(f_i\) for the evolution process.

\paragraph{Crossover}
During sexual reproduction, chromosomal crossover creates novel genetic combinations that increase population diversity, providing raw material for natural selection to enhance adaptability.
\ga draws on this mechanism by combining the weights of different models. Specifically, we first design a selection probability based on individual fitness: \( p_i = {f_i}/{\sum f_k} \), where $k=1,2,\dots, N$. Using this probability, we select parent pairs ($x_{p_1}$,$x_{p_2}$) with their corresponding weights ($\mathbf{w}_{p_1}$, $\mathbf{w}_{p_2}$). The offspring is generated by combining these weights: $\mathbf{w}_{child} = t\cdot \mathbf{w}_{p_1} + (1-t)\cdot \mathbf{w}_{p_2}$, where $t = {f_{p_1}}/({f_{p_1}+f_{p_2}})$ is the normalized weight. The crossover rate (\(c_r\)) controls the proportion of population undergoing this process, preserving model diversity while improving performance.

\paragraph{Mutation}
In biological evolution, gene mutations introduce random variations that maintain population diversity. Following this principle, GENOME applies mutation in two stages. First, each individual \(x_i\) is selected for mutation with probability \(im_r\). Then, for each selected model, a binary mask \(\mathbf{m}_{i}\) is generated with probability \(gm_r\) to determine which weight parameters will be mutated. The mutation is executed as follows:  
\begin{equation}
\mathbf{w}'_{i} \leftarrow \mathbf{w}_{i} + \mathbf{m}_{i} \cdot \mathbf{E}_{i} , \quad \mathbf{E} \sim \mathcal{N}(0, \sigma^2)  
\end{equation}
where the binary mask \(\mathbf{m}_{i}\) identifies the parameters to change, while the Gaussian noise \(\mathbf{E}_{i}\) and its standard deviation \(\sigma\) control the mutation intensity. The resulting mutated weights \(\mathbf{w}'_{i}\) form new individuals \(x'\) in the population, balancing the preservation of successful traits with the introduction of innovative variations.

\paragraph{Selection}
Selection in \ga serves to simulate natural selection pressure while maintaining a fixed population size $N$. After crossover and mutation operations, the selection process first retains the top \(\alpha N\) individuals with the highest fitness scores (\(\alpha \in (0,1)\)). Then, to restore the population size to $N$, individuals are selected from the entire current population with probabilities proportional to their fitness scores, where the selection probability for individual $i$ is calculated as $p_i=f_i/\sum_{k=1}^N f_k$. This process effectively maintains population stability while favoring better-performing individuals.

\begin{algorithm}[!t]
\small
\caption{\textbf{\ga}}
\label{alg: GA}
\begin{algorithmic}
\STATE {\bfseries Input: }Task $T$, fitness function $f$, LLM experts $\{x_i\}_{i=1}^n$
\STATE {\bfseries Hyperparameters}: population size $N$, crossover rate $c_r$, mutation rates $(im_r, gm_r)$, sigma $\sigma$, elite ratio $\alpha$, max iteration $K$
\STATE {\bfseries \textit{// Initialization}}
\STATE $P \leftarrow \{\mathbf{w}_i = t\mathbf{w}_a + (1-t) \mathbf{w}_b | t \sim U(0,1)\}_{i=1}^N$
\STATE evaluate fitness $f(x_i)$ for $x_i \in P$
\STATE $g \leftarrow \arg\max_{x \in P} f(x)$
\FOR{$iter = 1$ {\bfseries to} $K$}
    \STATE {\bfseries \textit{// Crossover}}
    \FOR{each pair selected with probability $c_r$}
        \STATE $p_i \leftarrow f_i/\sum f_k,\quad k=1,2,\cdots,N$
        \STATE select parents $(x_{p_1}, x_{p_2})$ according to $\{p_i\}^N_{i}$
        \STATE $\mathbf{w}_{child} \leftarrow t\mathbf{w}_{p_1}+(1-t)\mathbf{w}_{p_2}$, create $x_{child}$ by $\mathbf{w}_{child}$
        \STATE evaluate $x_{child}$ on $f$
    \ENDFOR
    \STATE {\bfseries \textit{// Mutation}}
    \FOR{each individual with probability $im_r$}
        \FOR{each weight $\omega_{ij}$ with probability $gm_r$}
            \STATE $\mathbf{w}'_{i} \leftarrow \mathbf{w}_{i} + \mathbf{m}_{i}\cdot \mathbf{E}_{i}$, create $x_i'$ by $\mathbf{w}_{i}$
            \STATE evaluate $x_{i}'$ on $f$
        \ENDFOR
    \ENDFOR
    \STATE {\bfseries \textit{// Selection}}
    \STATE $P \leftarrow \text{elite}(\alpha \cdot N) \cup \text{fitness-based selection}((1-\alpha)\cdot N)$ 
    \STATE $g \leftarrow \mathop{\arg\max}_{x \in P} f(x)$  
\ENDFOR

\STATE evaluate test fitness $f_{\text{test}}(\mathbf{x}_i)$ for $\mathbf{x}_i \in P$, update $\mathbf{g}$
\STATE {\bfseries return} best individual $g$
\end{algorithmic}
\end{algorithm}

\subsection{GENOME+}
GA primarily focuses on generational evolution at the genetic level. The evolutionary trajectories of intelligent species suggest that knowledge transmission and collaborative decision-making within social groups are significant factors in species evolution. 
This evolutionary mechanism extends beyond genetic factors, offering new perspectives on the evolution of LLM groups. When models are conceptualized as a biological population, their evolution encompasses not only the crossover and mutation of weight parameters but also the capabilities of experience transfer and group decision-making.
We introduce two new operations in the \ga to achieve these capabilities. 
Through \textbf{succession}, individuals can derive successful experiences from the historical evolutionary trajectory of the population and circumvent past failures. 
\textbf{Ensemble} integrates the reasoning capabilities of various models to achieve the emergence of collective intelligence.
Algorithm \ref{alg: hybrid} summarizes the population evolution process.

\paragraph{Succession}
Model Swarms \citep{feng2024model} demonstrates that Particle Swarm Optimization (PSO) can effectively optimize LLM groups by tracking personal best, global best, and global worst solutions. Building on this insight, we represent each individual's learning pattern as an experience vector \( \mathbf{e}_i \), which is randomly initialized with the same dimensionality as the LoRA weights \( \mathbf{w}_i \), and facilitate knowledge transfer within the population through experience updates.
The experience update for each individual integrates four sources: global best \( \mathbf{e}_g \) (the best experience in the population's evolution so far), current best \( \mathbf{e}_c \) (the latest successful pattern adapting to dynamic environmental changes), global worst \( \mathbf{e}_w \) (failed experience as a negative example), and self-experience \( \mathbf{e}_i \) (the individual's accumulated experience). The update follows:
\vspace{-0.4em}
\begin{equation}
\mathbf{e}_i \leftarrow \frac{1}{\mathbf{C}}[\phi_e\mathbf{e}_i + \phi_g(\mathbf{e}_g - \mathbf{e}_i) + \phi_c(\mathbf{e}_c - \mathbf{e}_i) - \phi_w(\mathbf{e}_w - \mathbf{e}_i)]
\end{equation}

where \( \phi_* \) are the weights for each source of experience, and \( \mathbf{C} = \phi_e + \phi_g + \phi_c + \phi_w \) is the normalization coefficient. The updated experience is then applied to adjust the model weights: \( \mathbf{w}_i \leftarrow \mathbf{w}_i + \lambda \mathbf{e}_i \), where \( \lambda \) is the learning rate for experience. This mechanism allows individuals to maintain their unique characteristics while also absorbing the collective wisdom of the group during the evolutionary process.

\paragraph{Ensemble}
\gsa introduces an ensemble operation during the inference phase, selecting the top-k individuals with the highest fitness from the evolved population and aggregating their prediction results.
This ensemble decision-making mechanism integrate the outputs of multiple high fitness individuals, while fully leveraging the advantages accumulated by the population during the evolutionary process. Through result aggregation in the inference phase, \gsa can achieve more robust predictive performance.

\begin{algorithm}[!htbp]
\small
\caption{\textbf{GENOME+}}
\label{alg: hybrid}
\begin{algorithmic}
\STATE {\bfseries Input: } Task $T$, fitness function $f$, LLM experts $\{\mathbf{x}_i\}_{i=1}^n$
\STATE {\bfseries Hyperparameters:} population size $N$, crossover rate $c_r$, mutation rates $(im_r, gm_r)$, sigma $\sigma$, elite ratio $\alpha$, learning weight $\phi_\ast$, learning rate $\lambda$, max iteration $K$

\STATE {\bfseries Population Initialization}
\FOR{$iter = 1$ {\bfseries to} $K$}
    \STATE {\bfseries Crossover, Mutation}
    \STATE {\bfseries \textit{// succession}}
    \FOR{$i=1$ {\bfseries to} $N$} 
    \STATE \footnotesize{$\mathbf{e}_i \leftarrow \frac{1}{\mathbf{C}}[\phi_e\mathbf{e}_i + \phi_g(\mathbf{e}_g - \mathbf{e}_i) + \phi_c(\mathbf{e}_c - \mathbf{e}_i) - \phi_w(\mathbf{e}_w - \mathbf{e}_i)]$}
    \STATE \( \mathbf{w}_i \leftarrow \mathbf{w}_i + \lambda e_i \), create $x_i$ by $\mathbf{w}_i$
    \ENDFOR
    \STATE {\bfseries Selection}
\ENDFOR
\STATE evaluate test fitness $f_{\text{test}}$ for $x_i \in P$, update $\mathbf{g}$\;
\STATE {\bfseries \textit{// ensemble}}
\STATE Select top-$k$ individuals based on validation fitness $f_{\text{valid}}$\;
\STATE Combine selected individuals into an ensemble $\mathbf{E}$\;

\STATE {\bfseries return} best individual $\mathbf{g}$ and ensemble $\mathbf{E}$
    
\end{algorithmic}
\end{algorithm}

\section{Experiments}
In the following discussion, \textbf{GENOME(+)} refers to GENOME and GENOME+ together.
Our experiments investigate the following key questions: 
(i) How effectively GENOME(+) adapt to a single task compared to existing methods? 
(ii) Can the framework maintain stable performance when handling multiple tasks simultaneously? 
(iii) Can the framework  generalize to unseen, held-out tasks? 
(iv) How well does our framework scale with increasing population size?
(v) How does hardware configuration affect time efficiency and overall performance?
(vi) What is the impact of each operator on overall performance?

\subsection{Experimental Setup}

\paragraph{Models} We employ \textit{gemma-2-2b-it} \citep{team2024gemma} as our foundation model, and construct a set of domain-specific experts by fine-tuning the foundation model on 10 distinct domains extracted from the \textbf{Tulu-v2-SFT-mixture} dataset \citep{ivison2023tuluv2}. The fine-tuning process is implemented using the llama-factory framework \citep{zheng2024llamafactory}, incorporating the low-rank adaptation (LoRA) technique \citep{hu2021lora}. 
We further demonstrate the capabilities of these 10 experts across different domains (see Figure \ref{fig: experts}), which confirms that our training process yields expert models with varying proficiencies. Complete training hyperparameters and configurations are detailed in Table \ref{tab: sft parameters}.
 


\paragraph{Datasets}
We consider 12 datasets covering 7 key capabilities of LLMs, including \textbf{General Knowledge} (MMLU \citep{hendryckstest2021}, MMLUPro \citep{wang2024mmlu}), \textbf{Mathematics} (MATH \citep{hendrycksmath2021}, GSM8K \citep{cobbe2021gsm8k}, 
and MGSM \citep{shi2022languagemodelsmultilingualchainofthought}), \textbf{Code Generation} (MBPP \citep{austin2021program}), 
\textbf{Logical Reasoning} (DROP \citep{dua2019dropreadingcomprehensionbenchmark}, BBH \citep{suzgun2022challenging}), 
\textbf{Multilingual Processing} (MGSM \citep{shi2022languagemodelsmultilingualchainofthought} and FLORES-101 \citep{flores101}), 
\textbf{Affective Computing} (EmoryNLP \citep{zahiri2018emotion}, MELD~\citep{poria2018meld}), and \textbf{Question Answering} (ARC\_C \citep{clark2018thinksolvedquestionanswering} and CSQA \citep{talmor-etal-2019-commonsenseqa}). 
Note that MGSM is a multilingual extension of mathematics problems, thus appearing in both Mathematics and Multilingual Processing.
Each dataset is divided into a 200-sample validation set and a test set comprising approximately 1,000 samples. Detailed descriptions of the data splits and metrics can be found in Appendix \ref{app: dataset} and Table \ref{tab: dataset}. Notably, these datasets do not overlap with the training data used for the expert models.

\begin{table*}[!htbp]
\small
\centering
\resizebox{0.8\textwidth}{!}{%
\begin{tabular}{lllcc}
\toprule
 &   &   & \multicolumn{2}{c}{\textbf{  Size}} \\ 
\cline{4-5} 
\multirow{-2}{*}{\textbf{Dataset}} & \multirow{-2}{*}{\textbf{Category}} & \multirow{-2}{*}{\textbf{Metrics}}                   & validation & test \\ 
\midrule
ARC\_C    & Question Answering     & accuracy, 0-shot    & 200        & 1000        \\
BBH       & Logical Reasoning           & accuracy, 3-shot    & 200        & 1000        \\
CSQA      & Question Answering     & accuracy, 0-shot    & 200        & 1000        \\
DROP      & Logical Reasoning           & exact match, 0-shot & 200        & 1000        \\
EmoryNLP  & Affective Computing & weighted-F1, 0-shot & 200        & 697         \\
Flores-37/101 & Multilingual Processing       & BLEU, 3-shot        & 200        & 1012        \\
GSM8k     & Mathematics                & accuracy, 0-shot    & 200        & 1000        \\
MATH      & Mathematics                & accuracy, 0-shot    & 200        & 1000        \\
MBPP      & Code Generation              & Pass@1, 0-shot      & 200        & 774  \\
MELD & Affective Computing & weighted-F1, 0-shot & 200 & 1000 \\
MGSM      & Multilingual Processing, Mathematics  & accuracy, 0-shot    & 200        & 2637        \\
MMLU      & General Knowledge   & accuracy, 0-shot    & 200        & 1000        \\
MMLUPro  & General Knowledge   & accuracy, 0-shot    & 200        & 1000        \\ \bottomrule
\end{tabular}%
}
\caption{Detailed information of the datasets.}
\label{tab: dataset}
\end{table*}

\paragraph{Implementation of GENOME(+)} 
Unless otherwise stated, we use the following setting: crossover rate ($c_r$) of 0.3, individual mutation rate ($im_r$) of 0.3, gene mutation rate ($gm_r$) of 0.2, and standard deviation ($\sigma$) of 0.001. Additionally, the maximum number of iterations is set to 10, the population size is 10, $\phi_\ast$ is (0.95, 0.2, 0.2, 0.1), $\lambda$ is 0.95.
For ensemble operation, we select the top-3 individuals with the highest fitness scores on the validation set.

\paragraph{Baselines}
We compare GENOME(+) to 7 baselines:
\begin{itemize}[leftmargin=*]
\setlength\itemsep{0em} 
\vspace{-0.8em}
    \item Best Single, which refers to the best-performing expert model for a given dataset.
    \item Data Merge, which combines the 10 sub-datasets of Tulu-v2-sft-mixture into one complete training dataset for training a single LoRA model.
     \item Expert Fusion, which merges the 10 expert models through \(\mathbf{w}_\text{fusion}=\sum_{i=1}^n t_i \mathbf{w}_i \), where $t_i$ denotes the normalized fitness value of each expert model $i$. 
    \item {DARE\_TIES} \citep{yu2024dareties}, which merges multiple LoRA models into a single one, while effectively reducing parameter interference.
    \item LoraHub \citep{huang2024lorahubefficientcrosstaskgeneralization}, which dynamically selecting and merging LoRA models, uses GA to optimize the combination weights.
    \item Pack of LLMs \citep{mavromatis2024packllmsmodelfusion}, which dynamically merges LoRA models, assigning weights based on each model's perplexity for given input prompts. 
    \item Model Swarms \citep{feng2024model}, which employs PSO to iteratively  optimize the set of expert models and dynamically adapt to a given dataset.
\end{itemize}
Note that Expert Fusion, LoraHub, Pack of LLMs, Model Swarms, and our framework all require a few samples for dynamic adaptation (\textit{dynamic methods}).
For a fair comparison, we fix the number of samples to 200 across all methods.
Additionally, all aforementioned methods---except for Data Merge, which does not require multiple expert models---utilize the same set of 10 expert models trained on the Tulu-v2-SFT-mixture.
The implementation details of these methods are illustrated in Appendix~\ref{sec: baseline detailed}.


\subsection{Results and Analysis}
Unless specified otherwise, all experiments are conducted using five random seeds, and the results are averaged.

\subsubsection{Single Task}
\begin{table*}[!t]
\centering
\resizebox{\textwidth}{!}{%
\begin{tabular}{@{}lcccccccccccc@{}}
\toprule
\textbf{Method} &
  \textbf{MMLU} &
  \textbf{MMLUPro} &
  \textbf{GSM8k} &
  \textbf{MATH} &
  \textbf{MGSM} &
  \textbf{Flores-37} &
  \textbf{ARC\_C} &
  \textbf{CSQA} &
  \textbf{BBH} &
  \textbf{DROP} &
  \textbf{EmoryNLP} &
  \textbf{MBPP} \\ \midrule
\textbf{Best Single} &
  52.90 &
  26.57 &
  40.80 &
  14.30 &
  30.22 &
  21.51 &
  57.34 &
  64.30 &
  30.22 &
  30.40 &
  32.53 &
  31.78 \\
\textbf{Data Merge} &
  13.60 &
  25.57 &
  33.00 &
  12.20 &
  25.98 &
  22.51 &
  46.08 &
  37.40 &
  25.98 &
  27.40 &
  32.37 &
  15.37 \\
\textbf{DARE\_TIES}  &
  48.30 &
  28.57 &
  35.50 &
  12.50 &
  31.85 &
  22.33 &
  69.80 &
  23.90 &
  31.85 &
  19.22 &
  34.22 &
  37.77 \\ \midrule
\textbf{Expert Fusion} &
  55.80 &
  27.67 &
  39.50 &
  13.10 &
  31.82 &
  21.74 &
  69.03 &
  65.00 &
  31.82 &
  22.60 &
  32.66 &
  28.55 \\
\textbf{LoraHub}  &
  53.00 &
  27.17 &
  46.47 &
  14.90 &
  34.70 &
  21.83 &
  69.97 &
  66.10 &
  39.30 &
  35.32 &
  31.53 &
  42.63 \\
\textbf{Pack of LLMs}  &
  54.30 &
  27.67 &
  38.13 &
  11.50 &
  31.06 &
  19.24 &
  70.22 &
  63.30 &
  38.90 &
  21.30 &
  30.46 &
  42.56 \\
\textbf{Model Swarms} &
  55.91 &
  27.77 &
  45.82 &
  15.06 &
  33.15 &
  21.76 &
  68.53 &
  \underline{68.76} &
  38.80 &
  34.52 &
  33.98 &
  42.56 \\ \midrule
\textbf{\ga} &
  \textbf{56.66} &
  \underline{27.77} &
  \underline{49.34} &
  \underline{15.72} &
  \underline{37.19} &
  \underline{22.78} &
  \underline{70.69} &
  68.22 &
  {\ul 40.00} &
  \underline{35.28} &
  \underline{34.97} &
  \textbf{43.60} \\
\textit{\textbf{ }\footnotesize{- vs Best Single}} &
  \footnotesize{\color[HTML]{FF0000} +7.11\%} &
  \footnotesize{\color[HTML]{FF0000} +4.52\%} &
  \footnotesize{\color[HTML]{FF0000} +20.93\%} &
  \footnotesize{\color[HTML]{FF0000} +9.93\%} &
  \footnotesize{\color[HTML]{FF0000} +23.06\%} &
  \footnotesize{\color[HTML]{FF0000} +5.90\%} &
  \footnotesize{\color[HTML]{FF0000} +23.29\%} &
  \footnotesize{\color[HTML]{FF0000} +6.10\%} &
  \footnotesize{\color[HTML]{FF0000} +32.36\%} &
  \footnotesize{\color[HTML]{FF0000} +16.05\%} &
  \footnotesize{\color[HTML]{FF0000} +7.50\%} &
  \footnotesize{\color[HTML]{FF0000} +37.19\%} \\
\textit{\textbf{ }\footnotesize{- vs Model Swarms}} &
  \footnotesize{\color[HTML]{FF0000} +1.34\%} &
  -- &
  \footnotesize{\color[HTML]{FF0000} +7.68\%} &
  \footnotesize{\color[HTML]{FF0000} +4.38\%} &
  \footnotesize{\color[HTML]{FF0000} +12.19\%} &
  \footnotesize{\color[HTML]{FF0000} +4.69\%} &
  \footnotesize{\color[HTML]{FF0000} +3.15\%} &
  \footnotesize{\color[HTML]{32CB00} -0.79\%} &
  \footnotesize{\color[HTML]{FF0000} +3.09\%} &
  \footnotesize{\color[HTML]{FF0000} +2.20\%} &
  \footnotesize{\color[HTML]{FF0000} +2.91\%} &
  \footnotesize{\color[HTML]{FF0000} +2.44\%} \\

\textbf{GENOME+} &
  \underline{56.44} &
  \textbf{30.98} &
  \textbf{51.24} &
  \textbf{16.41} &
  \textbf{39.55} &
  \textbf{23.38} &
  \textbf{74.38} &
  \textbf{69.89} &
  \textbf{41.10} &
  \textbf{47.06} &
  \textbf{38.81} &
  {\ul 43.54} \\
\footnotesize\textit{\textbf{ }- vs Best Single} &
  \footnotesize{\color[HTML]{FF0000} +6.69\%} &
  \footnotesize{\color[HTML]{FF0000} +16.60\%} &
  \footnotesize{\color[HTML]{FF0000} +25.59\%} &
  \footnotesize{\color[HTML]{FF0000} +14.76\%} &
  \footnotesize{\color[HTML]{FF0000} +30.87\%} &
  \footnotesize{\color[HTML]{FF0000} +8.69\%} &
  \footnotesize{\color[HTML]{FF0000} +29.72\%} &
  \footnotesize{\color[HTML]{FF0000} +8.69\%} &
  \footnotesize{\color[HTML]{FF0000} +36.00\%} &
  \footnotesize{\color[HTML]{FF0000} +54.80\%} &
  \footnotesize{\color[HTML]{FF0000} +19.31\%} &
  \footnotesize{\color[HTML]{FF0000} +37.00\%} \\
\footnotesize\textit{\textbf{ }- vs Model Swarms} &
  \footnotesize{\color[HTML]{FF0000} +0.95\%} &
  \footnotesize{\color[HTML]{FF0000} +11.56\%} &
  \footnotesize{\color[HTML]{FF0000} +11.83\%} &
  \footnotesize{\color[HTML]{FF0000} +8.96\%} &
  \footnotesize{\color[HTML]{FF0000} +19.31\%} &
  \footnotesize{\color[HTML]{FF0000} +7.44\%} &
  \footnotesize{\color[HTML]{FF0000} +8.54\%} &
  \footnotesize{\color[HTML]{FF0000} +1.64\%} &
  \footnotesize{\color[HTML]{FF0000} +5.93\%} &
  \footnotesize{\color[HTML]{FF0000} +36.33\%} &
  \footnotesize{\color[HTML]{FF0000} +14.21\%} &
  \footnotesize{\color[HTML]{FF0000} +2.30\%} \\ 
  \bottomrule
\end{tabular}%
}
\caption{Performance comparison of different methods across 12 datasets (averaged over 5 runs with different random seeds). Best results are in bold and second-best results are underlined. 
In comparison rows, {\color[HTML]{FF0000}red}/{\color[HTML]{32CB00}green} percentages indicate improvements/decreases respectively, and `—' denotes no change. \gsa achieves average improvements of \textbf{24.06\%/10.75\%} over Best Single/Model Swarms respectively (max \textbf{54.80\%/36.33\%} on DROP). }
\label{tab: single task}
\end{table*}

Table \ref{tab: single task} demonstrates that the \ga and \gsa have attained superior performance across \textbf{12 datasets}. Specifically, \gsa exhibits the most robust performance---on average, enhancing by 24.06\% relative to Best Single and by 10.75\% in comparison to Model Swarms, thoroughly substantiating its efficacy. Our frameworks have demonstrated consistent performance enhancements across multiple categories, particularly in tasks necessitating intricate reasoning skills, such as BBH and DROP, where the framework attain increases of 36\% and 54.80\%, respectively. In Mathematics, it increases by 23.74\% relative to the Best Single and by 13.37\% in comparison to Model Swarms. 

\begin{table}[!htbp]
\small
\centering
\begin{tabular}{@{}lll@{}}
\toprule
\textbf{Method} & \textbf{MMLUPro$_\text{Knowledge}$} & \textbf{MMLUPro$_\text{Reasoning}$} \\ \midrule
\textbf{Best Single} & 27.90 & 23.20 \\
\textbf{Model Swarms} & 27.36 {\footnotesize\color[HTML]{32CB00}(-1.94\%)} & 24.95 {\footnotesize\color[HTML]{FE0000}(+7.54\%)} \\
\textbf{\gaa} & 27.48 {\footnotesize\color[HTML]{FE0000}(+0.44\%)} & 25.24 {\footnotesize\color[HTML]{FE0000}(+8.79\%)} \\
\textbf{\gsaa} & 29.05 {\footnotesize\color[HTML]{FE0000}(+6.18\%)} & 28.73 {\footnotesize\color[HTML]{FE0000}(+23.84\%)} \\ \bottomrule
\end{tabular}
\vspace{0.2em}
\caption{Performance comparison of different  methods on MMLUPro knowledge and reasoning subsets. The percentages in parentheses indicate the relative improvement ({\color[HTML]{FE0000}red}) or degradation ({\color[HTML]{32CB00}green}) compared to the Best Single.}
\label{tab: mmlupro_sub}
\end{table}

Nonetheless, we notice that in Question Answering (ARC\_C, CSQA) and General Knowledge (MMLU, MMLUPro), the enhancements in performance are less significant.
To verify whether this difference is influenced by task type, we construct subsets of knowledge-based and reasoning-based tasks within the MMLUPro. The former includes tasks that rely on domain knowledge such as law, history, and psychology, while the latter comprises fields that require logical reasoning, such as mathematics, physics, and computer science. As shown in Table \ref{tab: mmlupro_sub}, the improvement rates of all methods on reasoning-based tasks are much higher than those on knowledge-based tasks. Notably, \gsa achieves a performance gain of 23.84\% on the reasoning subset compared to the Best Single, whereas the gain is  6.18\% on the knowledge subset. Other methods exhibit a similar trend. These results reveal that existing dynamic methods, including our population-based evolution framework, excels in reasoning-intensive tasks. On the other hand, knowledge-intensive tasks may require the integration of external knowledge bases or knowledge enhancement strategies, which is beyond the capability of dynamic methods.

\subsubsection{Multi-Task Domain}
In our second experiment, we consider five domains each consists of two tasks. To allow for simultaneous adapation to two task,s we consider the fitness value to be the average of the two tasks' performance. We only compare dynamic methods because the performance of static methods remains constant across tasks. Compared to single task, this scenario is more challenging.

As shown in Table \ref{tab: multi-task}, \ga and \gsa are still able to achieve performance at the level comparable to single-task scenario, whereas other dynamic methods suffer from various degrees of performance degradation. Interestingly, \gsa achieves the best results in every category, proving the stability of our framework in navigating challenging situations.

\begin{table*}[t]
\centering
\small
\resizebox{\textwidth}{!}{%
\begin{tabular}{lcccccccccc}
\toprule
\multirow{2.5}{*}{\textbf{Method}} &
  \multicolumn{2}{c}{\textbf{\footnotesize Affective Computing}} &
  \multicolumn{2}{c}{\textbf{\footnotesize Mathematics}} &
  \multicolumn{2}{c}{\textbf{\footnotesize General Knowledge}} &
  \multicolumn{2}{c}{\textbf{\footnotesize Question Answering}} &
  \multicolumn{2}{c}{\textbf{\footnotesize Logical Reasoning}}
  \\ 
  \cmidrule(lr){2-3} 
  \cmidrule(lr){4-5} 
  \cmidrule(lr){6-7} 
  \cmidrule(lr){8-9} 
  \cmidrule(lr){10-11} 
 &
  \textbf{\footnotesize MELD} &
  \textbf{\footnotesize EmoryNLP} &
  \textbf{\footnotesize GSM8k} &
  \textbf{\footnotesize MATH} &
  \textbf{\footnotesize MMLU} &
  \textbf{\footnotesize MMLUPro} &
  \textbf{\footnotesize ARC\_C} &
  \textbf{\footnotesize CSQA} & 
  \textbf{\footnotesize DROP} &
  \textbf{\footnotesize BBH} \\ \midrule
\textbf{Expert Fusion} & 50.35 & 31.91 & 41.77  & 12.60 & 53.10 & 25.97 & 68.94 & 63.00 & 22.10 & 38.30
\\
\textbf{LoraHub} & 50.77 & 32.67 & 41.77 & 10.30 & 52.90 & 26.37 & 69.37 & 62.40 & 34.10 & 38.30
\\
\textbf{Pack of LLMs} & 51.70 & 29.81 & 37.90 & 11.90 & 53.40 & 27.07 & 69.03 & 63.10 & 21.20 & 38.60 
\\
\textbf{Model Swarms}              & 52.31 & 33.53 & 43.84 & \underline{15.10} & 55.50 & 27.99 & 67.76 & 65.78 & 31.80 & 38.40 
\\ \midrule
\textbf{\gaa}       
& \underline{52.68} 
& \underline{35.08} 
& \underline{46.64} & 14.48 & \underline{55.52} & \underline{28.31} & \underline{69.73} & \underline{67.92} & \underline{37.90} & \underline{39.15} 
\\
\footnotesize\textit{\textbf{ }- vs Model Swarms} 
& \footnotesize{\color[HTML]{FE0000}+0.71\%} 
& \footnotesize{\color[HTML]{FE0000}+4.62\%} 
& \footnotesize{\color[HTML]{FE0000}+6.39\%} 
& \footnotesize{\color[HTML]{32CB00}-4.11\%}  
& \footnotesize{\color[HTML]{FE0000}+0.04\%} 
& \footnotesize{\color[HTML]{FE0000}+1.14\%}  
& \footnotesize{\color[HTML]{FE0000}+2.91\%} 
& \footnotesize{\color[HTML]{FE0000}+3.25\%} 
& \footnotesize{\color[HTML]{FE0000}+19.18\%} 
& \footnotesize{\color[HTML]{FE0000}+1.95\%} 
\\
\textbf{\gsaa} & \textbf{56.05} & \textbf{39.42} & \textbf{50.87} & \textbf{17.24} & \textbf{56.40} & \textbf{30.13} & \textbf{73.60} & \textbf{70.35} &  \textbf{48.00} & \textbf{39.90} 
\\ 
\footnotesize{\textit{\textbf{ }- vs Model Swarms}} 
& \footnotesize{\color[HTML]{FE0000}+7.15\%} 
& \footnotesize{\color[HTML]{FE0000}+17.57\%} 
& \footnotesize{\color[HTML]{FE0000}+16.04\%} 
& \footnotesize{\color[HTML]{FE0000}+14.17\%} 
& \footnotesize{\color[HTML]{FE0000}+1.62\%} 
& \footnotesize{\color[HTML]{FE0000}+7.65\%} 
& \footnotesize{\color[HTML]{FE0000}+8.62\%} 
& \footnotesize{\color[HTML]{FE0000}+6.95\%} 
& \footnotesize{\color[HTML]{FE0000}+50.94\%} 
& \footnotesize{\color[HTML]{FE0000}+3.91\%} 
\\
\bottomrule
\end{tabular}%
}

\caption{The results of multi-task domain.}
\label{tab: multi-task}
\end{table*}

\subsubsection{Zero-shot Generalization}
We assess the generalization capability of our framework through two experiments: cross-dataset transfer and unseen language adaptation.

We identify three groups of related task pairs for cross-dataset transfer: MMLUPro to MMLU, MATH to GSM8k, and EmoryNLP to MELD. For each pair, we initially enable a  dynamic method to adapt to one task with 200 samples, and then test its performance on the other task without any samples.  Table \ref{tab: zeroshot} illustrates that all dynamic methods exhibit generalization capabilities, with \ga and \gsa displaying the best performance. \gsa demonstrates a notable performance enhancement of 11.79\% compared to baseline methods.

\begin{table}[!h]
\small
\centering
\begin{tabular}{lccc}
\toprule
\textbf{Method} & \textbf{MMLUPro $\rightarrow$ MMLU} & \textbf{MATH $\rightarrow$ GSM8k} & \textbf{EmoryNLP $\rightarrow$ MELD}  \\
\midrule
\textbf{Expert Fusion}       & 53.40 & 42.53 & 50.75 \\
\textbf{LoraHub}             & 53.50 & 43.37    & 49.92       \\
\textbf{Pack of LLMs}        & 52.70 & 37.83   & 51.93       \\
\textbf{Model Swarms}                 & 52.10 & 44.05 & 51.29 \\
\textbf{\gaa} & \underline{55.40} & \underline{46.32} & \underline{52.07} \\
\textbf{\gsaa} & \textbf{56.65} & \textbf{51.10} & \textbf{55.28}  \\
\bottomrule
\end{tabular}
\vspace{0.2em}
\caption{Generalization performance across different domains}
\label{tab: zeroshot}
\end{table}

Additionally, we assess our framework using the Flores101 dataset, comprising 64 languages not included in the training phase of gemma-2\footnote{\href{https://cloud.google.com/vertex-ai/generative-ai/docs/learn/models}{gemma model support language list.}}. 
Table \ref{tab: multilingual} indicates that the majority of methods demonstrate performance improvements, with the exception of the Pack of LLMs. GENOME and GENOME+ demonstrate significant enhancements of 12.83\% and 15.98\%, respectively, exceeding their performance improvements on Flores37, which includes languages encountered during pre-training. The results indicate a robust generalization capability of our framework, especially in the context of low-resource languages.

\begin{table}[htbp]
\centering
\small
\begin{tabular}{@{}lll@{}}
\toprule
\textbf{Method} & \textbf{Flores-37} & \textbf{Flores-101} \\ \midrule
\textbf{Best Single} & 21.51 & 12.70 \\
\textbf{Expert Fusion} & 21.74 {\footnotesize\color[HTML]{FE0000}(+1.07\%)} & 13.74 {\footnotesize\color[HTML]{FE0000}(+8.19\%)}\\
\textbf{LoraHub} & 21.83 {\footnotesize\color[HTML]{FE0000}(+1.49\%)} & 13.82 {\footnotesize\color[HTML]{FE0000}(+8.82\%)}\\
\textbf{Pack of LLMs} & 19.24 {\footnotesize\color[HTML]{32CB00}(-10.55\%)} & 12.40 {\footnotesize\color[HTML]{32CB00}(-2.36\%)} \\
\textbf{Model Swarms} & 21.76 {\footnotesize\color[HTML]{FE0000}(+1.16\%)} & 13.64 {\footnotesize\color[HTML]{FE0000}(+7.40\%)} \\
\textbf{\gaa} & 22.78 {\footnotesize\color[HTML]{FE0000}(+5.90\%)} & 14.33 {\footnotesize\color[HTML]{FE0000}(+12.83\%)} \\
\textbf{\gsaa} & 23.38 {\footnotesize\color[HTML]{FE0000}(+8.69\%)} & 14.73 {\footnotesize\color[HTML]{FE0000}(+15.98\%)} \\ \bottomrule
\end{tabular}
\vspace{0.2em}
\caption{Performance comparison on seen (\textbf{Flores-37}) and unseen languages (\textbf{Flores-101}).}
\label{tab: multilingual}
\end{table}

\subsubsection{Scalability}
\begin{figure*}[!htbp]
\centering
\includegraphics[width=1.0\linewidth]{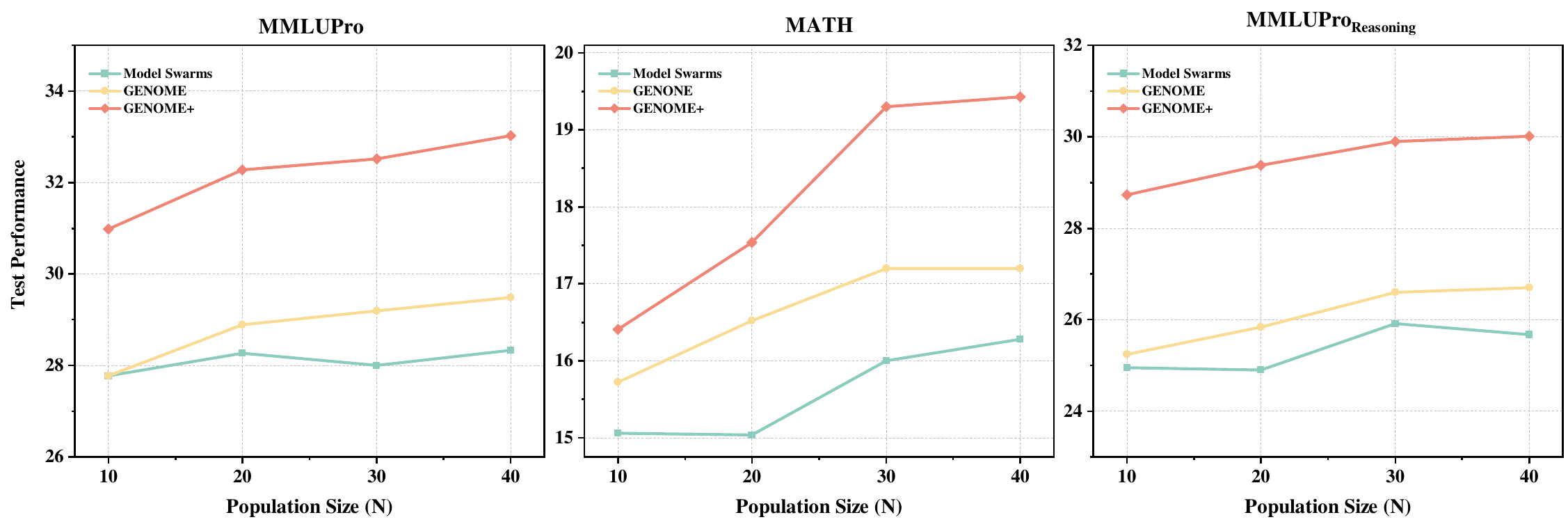}
\caption{Performance trends with increasing population sizes ($N$) across different methods.}
\label{fig: scaling}
\end{figure*}

We conduct comparison experiments using Model Swarms, \ga, and \gsa to examine their scalability, increasing the population size from 10 to 40 by 10. In the MMLUpro, MATH, and MMLUPro$_{\text{Reasoning}}$ tasks, all approaches exhibit enhanced performance with an increase in population size, with \gsa attaining the highest performance across all sizes. 
As shown in Figure \ref{fig: scaling}, \ga consistently outperforms Model Swarms, while \gsa demonstrates the strongest performance. These results confirm that \gsa can effectively scale to larger population sizes up to 40.

\subsubsection{Time Efficiency on Different GPUs}
To evaluate the time efficiency of our method, we conduct experiments on two different GPUs: \textbf{NVIDIA RTX 4090 (24GB VRAM)} and \textbf{NVIDIA A100 (80GB VRAM)}. 
All experiments run on a single GPU and are repeated five times to obtain the average result. The base model is gemma-2-2b-it, and the inference framework is vLLM. The GPU memory utilization is set to 95\% on both the A100 and 4090 to ensure a fair comparison. The population size set in 10, and other hyperparameters remain consistent with those used in the single-task experiments.

As shown in Table \ref{tab:speed_comparison}, the inference speed of the 4090 and A100 varies across different tasks. Despite having less VRAM, the 4090 maintains high inference performance on MMLUPro, DROP, and MATH tasks, further validating the applicability of our method.
It is important to note that \textbf{CPU performance}, \textbf{memory bandwidth}, and \textbf{system load} may also impact inference speed. Therefore, while our experimental results provide an overall trend, variations may occur in different environments.

\begin{table}[h]
    \centering
    \small
    \begin{tabular}{llcccc}
        \toprule
        \multirow{2.5}{*}{\textbf{Method}} & \multirow{2.5}{*}{\textbf{Task}} & \multicolumn{2}{c}{\textbf{Score}} & \multicolumn{2}{c}{\textbf{Time (s)}} \\
        \cmidrule(lr){3-4} \cmidrule(lr){5-6}
        &  & \textbf{4090} & \textbf{A100} & \textbf{4090} & \textbf{A100} \\
        \midrule
        \multirow{3}{*}{\textbf{\ga}}  
        & MMLUPro  & \textbf{28.97} & 28.67 & $\sim$3600 & $\sim$1600 \\
        & DROP     & \textbf{35.80} & 35.32 & $\sim$1000 & $\sim$800  \\
        & MATH     & 15.56 & \textbf{15.78} & $\sim$4000 & $\sim$2600 \\
        \midrule
        \multirow{3}{*}{\textbf{\gsa}}  
        & MMLUPro  & \textbf{31.47} & 30.98 & $\sim$5000 & $\sim$3000 \\
        & DROP     & 46.20 & \textbf{46.92} & $\sim$2000 & $\sim$1200 \\
        & MATH     & 16.79 & \textbf{16.81} & $\sim$6000 & $\sim$3800 \\
        \bottomrule
    \end{tabular}
    \vspace{0.2em}
    \caption{Comparison of time and score on different GPUs}
    \label{tab:speed_comparison}
\end{table}

\subsubsection{Ablation Study}
We perform ablation experiments on the MMLUPro, GSM8k, and DROP to confirm the contribution of each operation in the \gsa. As can be seen from Table \ref{tab: ablation}, the findings suggest that every operator of \gsa has a positive impact on performance. For comparison, we use \textit{random selection} in place of selection because eliminating it directly would result in an unlimited growth of the population size. Because random selection might prevent high-fitness individuals from being kept and selection operations impact the direction of population evolution, the data show an average performance loss of 14.68\%. For the ensemble operation, its removal results in an average performance drop of 11.27\%, especially a steep reduction in 21.38\% in the DROP. This suggests that the capacity of collective decision-making of multiple expert models is essential for improving task performance. With performance drops of 7.84\% and 7.94\%, respectively, the effects of the initialization and succession processes are similar, highlighting the significance of a high-quality starting population and efficient knowledge transfer in preserving population quality. The crossover and mutation operations are crucial for maintaining population diversity, despite their modest consequences (performance decreases of 3.65\% and 4.72\%, respectively).

\begin{table}[htbp]
\small
\centering
\begin{tabular}{llll}
\toprule
\textbf{Setting} & \textbf{MMLUPro} & \textbf{GSM8k} & \textbf{DROP} \\
\midrule
\textbf{\gsaa} & \textbf{30.98} & \textbf{51.24} & \textbf{47.06} \\
 \footnotesize\textit{w/o initialization} 
 & 29.72 {\footnotesize\color[HTML]{32CB00}(-4.07\%)} 
 & 46.02 {\footnotesize\color[HTML]{32CB00}(-10.19\%)} 
 & 42.70 {\footnotesize\color[HTML]{32CB00}(-9.26\%)} 
 \\
 \footnotesize\textit{w/o crossover} 
 & 30.32 {\footnotesize\color[HTML]{32CB00}(-2.13\%)} 
 & 48.86 {\footnotesize\color[HTML]{32CB00}(-4.64\%)} 
 & 45.10 {\footnotesize\color[HTML]{32CB00}(-4.16\%)} 
 \\
 \footnotesize\textit{w/o mutation} 
 & 29.72 {\footnotesize\color[HTML]{32CB00}(-4.07\%)} 
 & 49.51 {\footnotesize\color[HTML]{32CB00}(-3.38\%)} 
 & 43.90 {\footnotesize\color[HTML]{32CB00}(-6.71\%)} 
 \\
 \footnotesize\textit{random selection} 
 & 26.77 {\footnotesize\color[HTML]{32CB00}(-13.59\%)} 
 & 44.96 {\footnotesize\color[HTML]{32CB00}(-12.26\%)} 
 & 38.50 {\footnotesize\color[HTML]{32CB00}(-18.19\%)} 
 \\
 \footnotesize\textit{w/o succession} 
 & 28.76 {\footnotesize\color[HTML]{32CB00}(-7.17\%)}
 & 50.15 {\footnotesize\color[HTML]{32CB00}(-2.13\%)}
 & 40.22 {\footnotesize\color[HTML]{32CB00}(-14.53\%)}
 \\
 \footnotesize\textit{w/o ensemble} 
 & 28.81 {\footnotesize\color[HTML]{32CB00}(-7.00\%)}
 & 48.46 {\footnotesize\color[HTML]{32CB00}(-5.43\%)}
 & 37.00 {\footnotesize\color[HTML]{32CB00}(-21.38\%)}
 \\
\bottomrule
\end{tabular}
\vspace{0.2em}
\caption{Ablation study of different operations}
\label{tab: ablation}
\end{table}

\section{Related Work}
Existing approaches for combining LLMs fall into two main categories. One category adds extra components or routing mechanisms to coordinate multiple experts. A notable example is routing-based strategies~\citep{zhao2024loraretrieverinputawareloraretrieval,chen2024routerdcquerybasedrouterdual,shnitzer2023largelanguagemodelrouting}, but these methods demand extensive supervision and are not easily extended to new experts.
Another category focuses on parameter merging to harness complementary capabilities. Static methods (e.g., TIES~\citep{yadav2023resolving}, DARE \citep{yu2024dareties}) merge parameters without additional data but rely on predefined alignment rules, which limits their ability to generalize across diverse tasks. Dynamic methods employ extra data to guide the merging process and improve generalization. For instance, EvoMerge \citep{akibaEvolutionaryOptimizationModel2025} merges model weights via crossover (cross over), but it demands substantial supervision data and relies on backpropagation to update model parameters.

Some efforts (e.g., Model kinship \citep{hu2024exploring}, Pack of LLMs \citep{mavromatis2024packllmsmodelfusion}, Adamerging \citep{yang2024adamergingadaptivemodelmerging}) simplify the merging process using small amounts of data and simple metrics such as weight similarity or perplexity. Among these, LoRAHub \citep{huang2024lorahubefficientcrosstaskgeneralization} views model weights as a linear combination problem and leverages genetic algorithms to optimize the linear coefficients. Model Swarms \citep{feng2024model} extends this line of work by treating each LLM as a particle, applying particle swarm optimization (PSO) to obtain the best “particle.”

However, most of these methods center their merging process on eventually obtaining a single model that fuses multiple capabilities. While this approach is straightforward and does not require backpropagation, the inference phase then only retains the final merged model, thus losing the potential advantages of coordination among multiple models.

In contrast, our method adopts a “population” perspective: it maintains a stable population throughout the optimization process and makes full use of collective intelligence during inference. Algorithmically, compared with Model Swarms (which employs PSO), our method offers a more comprehensive evolutionary viewpoint. We treat model weights as “genes” and perform gene-level crossover, mutation, and selection, along with population-wide inheritance and ensemble operations.

Recently, there has been growing interest in applying evolutionary algorithms (EAs) to prompt optimization \citep{wuEvolutionaryComputationEra2024}, as illustrated by EmoPrompt \citep{emo-prompt-2024}, EoT \citep{jin2024zero}, EvoPrompt~\citep{guo2309connecting} and GPS \citep{xu2022gps}. EAs have demonstrated iterative improvements in model behavior for tasks such as emotional text generation and few-shot reasoning. Moreover, other studies have explored EAs for automated workflow (e.g., LLaMEA~\citep{llamea}, EvoAgent~\citep{yuan2024evoagentautomaticmultiagentgeneration}) design in LLMs. However, these studies primarily focus on optimizing prompts or workflows, with limited exploration of the comprehensive evolutionary adaptation of model weights themselves.

\section{Conclusion}
This paper formally defines the population-based evolution problem for LLMs and  presents two novel solutions: GENOME and GENOME+. These two frameworks leverage principles of biological evolution to adapt LLMs to new tasks with the use of few  samples.
Our experiments indicate that GENOME+ consistently surpasses existing methods across various tasks, showing notable enhancements in reasoning-intensive tasks. The framework demonstrates strong multi-task domain capabilities and effective generalization to novel tasks, while ensuring scalability with larger population sizes (up to 40). Ablation studies demonstrate the essential role of each evolutionary component, with selection and ensemble operations exhibiting the greatest influence on performance.
This study illustrates that population-based evolution presents a promising approach to enhance LLMs' capabilities through evolutionary  optimization. 
In addition, our experiments confirm that GENOME(+) can operate on devices equipped with 24GB of VRAM. We open-source our code and release the expert models, thereby promoting reproducibility and further research in this area. 
Future research may examine the integration of external knowledge bases and the exploration of advanced evolutionary mechanisms to enhance model adaptation capabilities.






\bibliography{reference.bib}

\begin{thebibliography}{49}
\expandafter\ifx\csname natexlab\endcsname\relax\def\natexlab#1{#1}\fi
\expandafter\ifx\csname url\endcsname\relax
  \def\url#1{\texttt{#1}}\fi
\expandafter\ifx\csname urlprefix\endcsname\relax\def\urlprefix{URL }\fi

\bibitem[{Akiba et~al.(2024{\natexlab{a}})Akiba, Shing, Tang, Sun and Ha}]{akiba2024evolutionary}
\textsc{Akiba, T.}, \textsc{Shing, M.}, \textsc{Tang, Y.}, \textsc{Sun, Q.} and \textsc{Ha, D.} (2024{\natexlab{a}}).
\newblock Evolutionary optimization of model merging recipes.
\newblock \textit{arXiv preprint arXiv:2403.13187} .

\bibitem[{Akiba et~al.(2024{\natexlab{b}})Akiba, Shing, Tang, Sun and Ha}]{akibaEvolutionaryOptimizationModel2025}
\textsc{Akiba, T.}, \textsc{Shing, M.}, \textsc{Tang, Y.}, \textsc{Sun, Q.} and \textsc{Ha, D.} (2024{\natexlab{b}}).
\newblock Evolutionary optimization of model merging recipes.
\newblock \textit{Nature Machine Intelligence} .

\bibitem[{Austin et~al.(2021)Austin, Odena, Nye, Bosma, Michalewski, Dohan, Jiang, Cai, Terry, Le et~al.}]{austin2021program}
\textsc{Austin, J.}, \textsc{Odena, A.}, \textsc{Nye, M.}, \textsc{Bosma, M.}, \textsc{Michalewski, H.}, \textsc{Dohan, D.}, \textsc{Jiang, E.}, \textsc{Cai, C.}, \textsc{Terry, M.}, \textsc{Le, Q.} \textsc{et~al.} (2021).
\newblock Program synthesis with large language models.
\newblock \textit{arXiv preprint arXiv:2108.07732} .

\bibitem[{Baumann and Kramer(2024)}]{emo-prompt-2024}
\textsc{Baumann, J.} and \textsc{Kramer, O.} (2024).
\newblock Evolutionary multi-objective optimization of large language model prompts for balancing sentiments.
\newblock In \textit{Applications of Evolutionary Computation} (S.~Smith, J.~Correia and C.~Cintrano, eds.). Springer Nature Switzerland, Cham.

\bibitem[{Chaudhary(2023)}]{codealpaca}
\textsc{Chaudhary, S.} (2023).
\newblock Code alpaca: An instruction-following llama model for code generation.
\newblock \url{https://github.com/sahil280114/codealpaca}.

\bibitem[{Chen et~al.(2024)Chen, Jiang, Lin, Kwok and Zhang}]{chen2024routerdcquerybasedrouterdual}
\textsc{Chen, S.}, \textsc{Jiang, W.}, \textsc{Lin, B.}, \textsc{Kwok, J.~T.} and \textsc{Zhang, Y.} (2024).
\newblock Routerdc: Query-based router by dual contrastive learning for assembling large language models.

\bibitem[{Chung et~al.(2024)Chung, Hou, Longpre, Zoph, Tay, Fedus, Li, Wang, Dehghani, Brahma et~al.}]{chung2024scaling}
\textsc{Chung, H.~W.}, \textsc{Hou, L.}, \textsc{Longpre, S.}, \textsc{Zoph, B.}, \textsc{Tay, Y.}, \textsc{Fedus, W.}, \textsc{Li, Y.}, \textsc{Wang, X.}, \textsc{Dehghani, M.}, \textsc{Brahma, S.} \textsc{et~al.} (2024).
\newblock Scaling instruction-finetuned language models.
\newblock \textit{Journal of Machine Learning Research} \textbf{25} 1--53.

\bibitem[{Clark et~al.(2018)Clark, Cowhey, Etzioni, Khot, Sabharwal, Schoenick and Tafjord}]{clark2018thinksolvedquestionanswering}
\textsc{Clark, P.}, \textsc{Cowhey, I.}, \textsc{Etzioni, O.}, \textsc{Khot, T.}, \textsc{Sabharwal, A.}, \textsc{Schoenick, C.} and \textsc{Tafjord, O.} (2018).
\newblock Think you have solved question answering? try arc, the ai2 reasoning challenge.

\bibitem[{Cobbe et~al.(2021)Cobbe, Kosaraju, Bavarian, Chen, Jun, Kaiser, Plappert, Tworek, Hilton, Nakano, Hesse and Schulman}]{cobbe2021gsm8k}
\textsc{Cobbe, K.}, \textsc{Kosaraju, V.}, \textsc{Bavarian, M.}, \textsc{Chen, M.}, \textsc{Jun, H.}, \textsc{Kaiser, L.}, \textsc{Plappert, M.}, \textsc{Tworek, J.}, \textsc{Hilton, J.}, \textsc{Nakano, R.}, \textsc{Hesse, C.} and \textsc{Schulman, J.} (2021).
\newblock Training verifiers to solve math word problems.
\newblock \textit{arXiv preprint arXiv:2110.14168} .

\bibitem[{Dua et~al.(2019)Dua, Wang, Dasigi, Stanovsky, Singh and Gardner}]{dua2019dropreadingcomprehensionbenchmark}
\textsc{Dua, D.}, \textsc{Wang, Y.}, \textsc{Dasigi, P.}, \textsc{Stanovsky, G.}, \textsc{Singh, S.} and \textsc{Gardner, M.} (2019).
\newblock Drop: A reading comprehension benchmark requiring discrete reasoning over paragraphs.

\bibitem[{Feng et~al.(2024)Feng, Wang, Wang, Ebrahimi, Palangi, Miculicich, Kulshrestha, Rauschmayr, Choi, Tsvetkov et~al.}]{feng2024model}
\textsc{Feng, S.}, \textsc{Wang, Z.}, \textsc{Wang, Y.}, \textsc{Ebrahimi, S.}, \textsc{Palangi, H.}, \textsc{Miculicich, L.}, \textsc{Kulshrestha, A.}, \textsc{Rauschmayr, N.}, \textsc{Choi, Y.}, \textsc{Tsvetkov, Y.} \textsc{et~al.} (2024).
\newblock Model swarms: Collaborative search to adapt llm experts via swarm intelligence.
\newblock \textit{arXiv preprint arXiv:2410.11163} .

\bibitem[{Goddard et~al.(2024)Goddard, Siriwardhana, Ehghaghi, Meyers, Karpukhin, Benedict, McQuade and Solawetz}]{goddard2024arcee}
\textsc{Goddard, C.}, \textsc{Siriwardhana, S.}, \textsc{Ehghaghi, M.}, \textsc{Meyers, L.}, \textsc{Karpukhin, V.}, \textsc{Benedict, B.}, \textsc{McQuade, M.} and \textsc{Solawetz, J.} (2024).
\newblock Arcee's mergekit: A toolkit for merging large language models.
\newblock \textit{arXiv preprint arXiv:2403.13257} .

\bibitem[{Goyal et~al.(2022)Goyal, Gao, Chaudhary, Chen, Wenzek, Ju, Krishnan, Ranzato, Guzmán and Fan}]{flores101}
\textsc{Goyal, N.}, \textsc{Gao, C.}, \textsc{Chaudhary, V.}, \textsc{Chen, P.-J.}, \textsc{Wenzek, G.}, \textsc{Ju, D.}, \textsc{Krishnan, S.}, \textsc{Ranzato, M.}, \textsc{Guzmán, F.} and \textsc{Fan, A.} (2022).
\newblock The flores-101 evaluation benchmark for low-resource and multilingual machine translation.
\newblock \textit{Transactions of the Association for Computational Linguistics} \textbf{10} 522--538.

\bibitem[{Guo et~al.(2023)Guo, Wang, Guo, Li, Song, Tan, Liu, Bian and Yang}]{guo2309connecting}
\textsc{Guo, Q.}, \textsc{Wang, R.}, \textsc{Guo, J.}, \textsc{Li, B.}, \textsc{Song, K.}, \textsc{Tan, X.}, \textsc{Liu, G.}, \textsc{Bian, J.} and \textsc{Yang, Y.} (2023).
\newblock Connecting large language models with evolutionary algorithms yields powerful prompt optimizers. arxiv 2023.
\newblock \textit{arXiv preprint arXiv:2309.08532} .

\bibitem[{Hendrycks et~al.(2021{\natexlab{a}})Hendrycks, Burns, Basart, Zou, Mazeika, Song and Steinhardt}]{hendryckstest2021}
\textsc{Hendrycks, D.}, \textsc{Burns, C.}, \textsc{Basart, S.}, \textsc{Zou, A.}, \textsc{Mazeika, M.}, \textsc{Song, D.} and \textsc{Steinhardt, J.} (2021{\natexlab{a}}).
\newblock Measuring massive multitask language understanding.
\newblock \textit{Proceedings of the International Conference on Learning Representations (ICLR)} .

\bibitem[{Hendrycks et~al.(2021{\natexlab{b}})Hendrycks, Burns, Kadavath, Arora, Basart, Tang, Song and Steinhardt}]{hendrycksmath2021}
\textsc{Hendrycks, D.}, \textsc{Burns, C.}, \textsc{Kadavath, S.}, \textsc{Arora, A.}, \textsc{Basart, S.}, \textsc{Tang, E.}, \textsc{Song, D.} and \textsc{Steinhardt, J.} (2021{\natexlab{b}}).
\newblock Measuring mathematical problem solving with the math dataset.
\newblock \textit{NeurIPS} .

\bibitem[{Holland(1992)}]{holland1992genetic}
\textsc{Holland, J.~H.} (1992).
\newblock Genetic algorithms.
\newblock \textit{Scientific american} \textbf{267} 66--73.

\bibitem[{Hu et~al.(2021)Hu, Shen, Wallis, Allen-Zhu, Li, Wang, Wang and Chen}]{hu2021lora}
\textsc{Hu, E.~J.}, \textsc{Shen, Y.}, \textsc{Wallis, P.}, \textsc{Allen-Zhu, Z.}, \textsc{Li, Y.}, \textsc{Wang, S.}, \textsc{Wang, L.} and \textsc{Chen, W.} (2021).
\newblock Lora: Low-rank adaptation of large language models.
\newblock \textit{arXiv preprint arXiv:2106.09685} .

\bibitem[{Hu et~al.(2024)Hu, Yao, Zhang, Deng and Chen}]{hu2024exploring}
\textsc{Hu, Y.}, \textsc{Yao, Y.}, \textsc{Zhang, N.}, \textsc{Deng, S.} and \textsc{Chen, H.} (2024).
\newblock Exploring model kinship for merging large language models.
\newblock \textit{arXiv preprint arXiv:2410.12613} .

\bibitem[{Huang et~al.(2024)Huang, Liu, Lin, Pang, Du and Lin}]{huang2024lorahubefficientcrosstaskgeneralization}
\textsc{Huang, C.}, \textsc{Liu, Q.}, \textsc{Lin, B.~Y.}, \textsc{Pang, T.}, \textsc{Du, C.} and \textsc{Lin, M.} (2024).
\newblock Lorahub: Efficient cross-task generalization via dynamic lora composition.

\bibitem[{Ivison et~al.(2023{\natexlab{a}})Ivison, Wang, Pyatkin, Lambert, Peters, Dasigi, Jang, Wadden, Smith, Beltagy et~al.}]{ivison2023tuluv2}
\textsc{Ivison, H.}, \textsc{Wang, Y.}, \textsc{Pyatkin, V.}, \textsc{Lambert, N.}, \textsc{Peters, M.}, \textsc{Dasigi, P.}, \textsc{Jang, J.}, \textsc{Wadden, D.}, \textsc{Smith, N.~A.}, \textsc{Beltagy, I.} \textsc{et~al.} (2023{\natexlab{a}}).
\newblock Camels in a changing climate: Enhancing lm adaptation with tulu 2.
\newblock \textit{arXiv preprint arXiv:2311.10702} .

\bibitem[{Ivison et~al.(2023{\natexlab{b}})Ivison, Wang, Pyatkin, Lambert, Peters, Dasigi, Jang, Wadden, Smith, Beltagy et~al.}]{ivison2023camels}
\textsc{Ivison, H.}, \textsc{Wang, Y.}, \textsc{Pyatkin, V.}, \textsc{Lambert, N.}, \textsc{Peters, M.}, \textsc{Dasigi, P.}, \textsc{Jang, J.}, \textsc{Wadden, D.}, \textsc{Smith, N.~A.}, \textsc{Beltagy, I.} \textsc{et~al.} (2023{\natexlab{b}}).
\newblock Camels in a changing climate: Enhancing lm adaptation with tulu 2.
\newblock \textit{arXiv preprint arXiv:2311.10702} .

\bibitem[{Jin et~al.(2024)Jin, Liu and Tan}]{jin2024zero}
\textsc{Jin, F.}, \textsc{Liu, Y.} and \textsc{Tan, Y.} (2024).
\newblock Zero-shot chain-of-thought reasoning guided by evolutionary algorithms in large language models.
\newblock \textit{arXiv preprint arXiv:2402.05376} .

\bibitem[{K{\"o}pf et~al.(2024)K{\"o}pf, Kilcher, von R{\"u}tte, Anagnostidis, Tam, Stevens, Barhoum, Nguyen, Stanley, Nagyfi et~al.}]{kopf2024openassistant}
\textsc{K{\"o}pf, A.}, \textsc{Kilcher, Y.}, \textsc{von R{\"u}tte, D.}, \textsc{Anagnostidis, S.}, \textsc{Tam, Z.~R.}, \textsc{Stevens, K.}, \textsc{Barhoum, A.}, \textsc{Nguyen, D.}, \textsc{Stanley, O.}, \textsc{Nagyfi, R.} \textsc{et~al.} (2024).
\newblock Openassistant conversations-democratizing large language model alignment.
\newblock \textit{Advances in Neural Information Processing Systems} \textbf{36}.

\bibitem[{Mavromatis et~al.(2024)Mavromatis, Karypis and Karypis}]{mavromatis2024packllmsmodelfusion}
\textsc{Mavromatis, C.}, \textsc{Karypis, P.} and \textsc{Karypis, G.} (2024).
\newblock Pack of llms: Model fusion at test-time via perplexity optimization.

\bibitem[{Mukherjee et~al.(2023)Mukherjee, Mitra, Jawahar, Agarwal, Palangi and Awadallah}]{mukherjee2023orca}
\textsc{Mukherjee, S.}, \textsc{Mitra, A.}, \textsc{Jawahar, G.}, \textsc{Agarwal, S.}, \textsc{Palangi, H.} and \textsc{Awadallah, A.} (2023).
\newblock Orca: Progressive learning from complex explanation traces of gpt-4.

\bibitem[{Muqeeth et~al.(2024)Muqeeth, Liu, Liu and Raffel}]{muqeeth2024learning}
\textsc{Muqeeth, M.}, \textsc{Liu, H.}, \textsc{Liu, Y.} and \textsc{Raffel, C.} (2024).
\newblock Learning to route among specialized experts for zero-shot generalization.
\newblock \textit{arXiv preprint arXiv:2402.05859} .

\bibitem[{Peng et~al.(2023)Peng, Li, He, Galley and Gao}]{peng2023instruction}
\textsc{Peng, B.}, \textsc{Li, C.}, \textsc{He, P.}, \textsc{Galley, M.} and \textsc{Gao, J.} (2023).
\newblock Instruction tuning with gpt-4.
\newblock \textit{arXiv preprint arXiv:2304.03277} .

\bibitem[{Poria et~al.(2018)Poria, Hazarika, Majumder, Naik, Cambria and Mihalcea}]{poria2018meld}
\textsc{Poria, S.}, \textsc{Hazarika, D.}, \textsc{Majumder, N.}, \textsc{Naik, G.}, \textsc{Cambria, E.} and \textsc{Mihalcea, R.} (2018).
\newblock Meld: A multimodal multi-party dataset for emotion recognition in conversations.
\newblock \textit{arXiv preprint arXiv:1810.02508} .

\bibitem[{Shi et~al.(2022)Shi, Suzgun, Freitag, Wang, Srivats, Vosoughi, Chung, Tay, Ruder, Zhou, Das and Wei}]{shi2022languagemodelsmultilingualchainofthought}
\textsc{Shi, F.}, \textsc{Suzgun, M.}, \textsc{Freitag, M.}, \textsc{Wang, X.}, \textsc{Srivats, S.}, \textsc{Vosoughi, S.}, \textsc{Chung, H.~W.}, \textsc{Tay, Y.}, \textsc{Ruder, S.}, \textsc{Zhou, D.}, \textsc{Das, D.} and \textsc{Wei, J.} (2022).
\newblock Language models are multilingual chain-of-thought reasoners.

\bibitem[{Shnitzer et~al.(2023)Shnitzer, Ou, Silva, Soule, Sun, Solomon, Thompson and Yurochkin}]{shnitzer2023largelanguagemodelrouting}
\textsc{Shnitzer, T.}, \textsc{Ou, A.}, \textsc{Silva, M.}, \textsc{Soule, K.}, \textsc{Sun, Y.}, \textsc{Solomon, J.}, \textsc{Thompson, N.} and \textsc{Yurochkin, M.} (2023).
\newblock Large language model routing with benchmark datasets.

\bibitem[{Suzgun et~al.(2022)Suzgun, Scales, Sch{\"a}rli, Gehrmann, Tay, Chung, Chowdhery, Le, Chi, Zhou,  and Wei}]{suzgun2022challenging}
\textsc{Suzgun, M.}, \textsc{Scales, N.}, \textsc{Sch{\"a}rli, N.}, \textsc{Gehrmann, S.}, \textsc{Tay, Y.}, \textsc{Chung, H.~W.}, \textsc{Chowdhery, A.}, \textsc{Le, Q.~V.}, \textsc{Chi, E.~H.}, \textsc{Zhou, D.},  and \textsc{Wei, J.} (2022).
\newblock Challenging big-bench tasks and whether chain-of-thought can solve them.
\newblock \textit{arXiv preprint arXiv:2210.09261} .

\bibitem[{Talmor et~al.(2019)Talmor, Herzig, Lourie and Berant}]{talmor-etal-2019-commonsenseqa}
\textsc{Talmor, A.}, \textsc{Herzig, J.}, \textsc{Lourie, N.} and \textsc{Berant, J.} (2019).
\newblock {C}ommonsense{QA}: A question answering challenge targeting commonsense knowledge.
\newblock In \textit{Proceedings of the 2019 Conference of the North {A}merican Chapter of the Association for Computational Linguistics: Human Language Technologies, Volume 1 (Long and Short Papers)}. Association for Computational Linguistics, Minneapolis, Minnesota.

\bibitem[{Team et~al.(2024)Team, Riviere, Pathak, Sessa, Hardin, Bhupatiraju, Hussenot, Mesnard, Shahriari, Ram{\'e} et~al.}]{team2024gemma}
\textsc{Team, G.}, \textsc{Riviere, M.}, \textsc{Pathak, S.}, \textsc{Sessa, P.~G.}, \textsc{Hardin, C.}, \textsc{Bhupatiraju, S.}, \textsc{Hussenot, L.}, \textsc{Mesnard, T.}, \textsc{Shahriari, B.}, \textsc{Ram{\'e}, A.} \textsc{et~al.} (2024).
\newblock Gemma 2: Improving open language models at a practical size.
\newblock \textit{arXiv preprint arXiv:2408.00118} .

\bibitem[{van Stein and Bäck(2024)}]{llamea}
\textsc{van Stein, N.} and \textsc{Bäck, T.} (2024).
\newblock Llamea: A large language model evolutionary algorithm for automatically generating metaheuristics.
\newblock \textit{IEEE Transactions on Evolutionary Computation}  1--1.

\bibitem[{Wang et~al.(2024)Wang, Ma, Zhang, Ni, Chandra, Guo, Ren, Arulraj, He, Jiang et~al.}]{wang2024mmlu}
\textsc{Wang, Y.}, \textsc{Ma, X.}, \textsc{Zhang, G.}, \textsc{Ni, Y.}, \textsc{Chandra, A.}, \textsc{Guo, S.}, \textsc{Ren, W.}, \textsc{Arulraj, A.}, \textsc{He, X.}, \textsc{Jiang, Z.} \textsc{et~al.} (2024).
\newblock Mmlu-pro: A more robust and challenging multi-task language understanding benchmark.
\newblock \textit{arXiv preprint arXiv:2406.01574} .

\bibitem[{Wu et~al.(2024)Wu, Wu, Wu, Feng and Tan}]{wuEvolutionaryComputationEra2024}
\textsc{Wu, X.}, \textsc{Wu, S.-h.}, \textsc{Wu, J.}, \textsc{Feng, L.} and \textsc{Tan, K.~C.} (2024).
\newblock Evolutionary computation in the era of large language model: Survey and roadmap.

\bibitem[{Xu et~al.(2023)Xu, Sun, Zheng, Geng, Zhao, Feng, Tao and Jiang}]{xu2023wizardlm}
\textsc{Xu, C.}, \textsc{Sun, Q.}, \textsc{Zheng, K.}, \textsc{Geng, X.}, \textsc{Zhao, P.}, \textsc{Feng, J.}, \textsc{Tao, C.} and \textsc{Jiang, D.} (2023).
\newblock Wizardlm: Empowering large language models to follow complex instructions.
\newblock \textit{arXiv preprint arXiv:2304.12244} .

\bibitem[{Xu et~al.(2022)Xu, Chen, Du, Shao, Wang, Li and Yang}]{xu2022gps}
\textsc{Xu, H.}, \textsc{Chen, Y.}, \textsc{Du, Y.}, \textsc{Shao, N.}, \textsc{Wang, Y.}, \textsc{Li, H.} and \textsc{Yang, Z.} (2022).
\newblock Gps: Genetic prompt search for efficient few-shot learning.
\newblock \textit{arXiv preprint arXiv:2210.17041} .

\bibitem[{Yadav et~al.(2023)Yadav, Tam, Choshen, Raffel and Bansal}]{yadav2023resolving}
\textsc{Yadav, P.}, \textsc{Tam, D.}, \textsc{Choshen, L.}, \textsc{Raffel, C.} and \textsc{Bansal, M.} (2023).
\newblock Resolving interference when merging models.
\newblock \textit{arXiv preprint arXiv:2306.01708} \textbf{1}.

\bibitem[{Yadav et~al.(2024)Yadav, Vu, Lai, Chronopoulou, Faruqui, Bansal and Munkhdalai}]{yadav2024matters}
\textsc{Yadav, P.}, \textsc{Vu, T.}, \textsc{Lai, J.}, \textsc{Chronopoulou, A.}, \textsc{Faruqui, M.}, \textsc{Bansal, M.} and \textsc{Munkhdalai, T.} (2024).
\newblock What matters for model merging at scale?
\newblock \textit{arXiv preprint arXiv:2410.03617} .

\bibitem[{Yang et~al.(2024)Yang, Wang, Shen, Liu, Guo, Wang and Tao}]{yang2024adamergingadaptivemodelmerging}
\textsc{Yang, E.}, \textsc{Wang, Z.}, \textsc{Shen, L.}, \textsc{Liu, S.}, \textsc{Guo, G.}, \textsc{Wang, X.} and \textsc{Tao, D.} (2024).
\newblock Adamerging: Adaptive model merging for multi-task learning.

\bibitem[{Yu et~al.(2024)Yu, Yu, Yu, Huang and Li}]{yu2024dareties}
\textsc{Yu, L.}, \textsc{Yu, B.}, \textsc{Yu, H.}, \textsc{Huang, F.} and \textsc{Li, Y.} (2024).
\newblock Language models are super mario: Absorbing abilities from homologous models as a free lunch.
\newblock In \textit{Forty-first International Conference on Machine Learning}.

\bibitem[{Yuan et~al.(2024)Yuan, Song, Chen, Tan, Li and Yang}]{yuan2024evoagentautomaticmultiagentgeneration}
\textsc{Yuan, S.}, \textsc{Song, K.}, \textsc{Chen, J.}, \textsc{Tan, X.}, \textsc{Li, D.} and \textsc{Yang, D.} (2024).
\newblock Evoagent: Towards automatic multi-agent generation via evolutionary algorithms.

\bibitem[{Zahiri and Choi(2018)}]{zahiri2018emotion}
\textsc{Zahiri, S.~M.} and \textsc{Choi, J.~D.} (2018).
\newblock Emotion detection on tv show transcripts with sequence-based convolutional neural networks.
\newblock In \textit{Workshops at the thirty-second aaai conference on artificial intelligence}.

\bibitem[{Zhang et~al.(2019)Zhang, Kishore, Wu, Weinberger and Artzi}]{zhang2019bertscore}
\textsc{Zhang, T.}, \textsc{Kishore, V.}, \textsc{Wu, F.}, \textsc{Weinberger, K.~Q.} and \textsc{Artzi, Y.} (2019).
\newblock Bertscore: Evaluating text generation with bert.
\newblock \textit{arXiv preprint arXiv:1904.09675} .

\bibitem[{Zhao et~al.(2024)Zhao, Gan, Wang, Zhou, Yang, Kuang and Wu}]{zhao2024loraretrieverinputawareloraretrieval}
\textsc{Zhao, Z.}, \textsc{Gan, L.}, \textsc{Wang, G.}, \textsc{Zhou, W.}, \textsc{Yang, H.}, \textsc{Kuang, K.} and \textsc{Wu, F.} (2024).
\newblock Loraretriever: Input-aware lora retrieval and composition for mixed tasks in the wild.

\bibitem[{Zheng et~al.(2024)Zheng, Zhang, Zhang, Ye, Luo, Feng and Ma}]{zheng2024llamafactory}
\textsc{Zheng, Y.}, \textsc{Zhang, R.}, \textsc{Zhang, J.}, \textsc{Ye, Y.}, \textsc{Luo, Z.}, \textsc{Feng, Z.} and \textsc{Ma, Y.} (2024).
\newblock Llamafactory: Unified efficient fine-tuning of 100+ language models.
\newblock In \textit{Proceedings of the 62nd Annual Meeting of the Association for Computational Linguistics (Volume 3: System Demonstrations)}. Association for Computational Linguistics, Bangkok, Thailand.

\bibitem[{Zhou et~al.(2023)Zhou, Liu, Xu, Iyer, Sun, Mao, Ma, Efrat, Yu, YU, Zhang, Ghosh, Lewis, Zettlemoyer and Levy}]{NEURIPS2023_lima}
\textsc{Zhou, C.}, \textsc{Liu, P.}, \textsc{Xu, P.}, \textsc{Iyer, S.}, \textsc{Sun, J.}, \textsc{Mao, Y.}, \textsc{Ma, X.}, \textsc{Efrat, A.}, \textsc{Yu, P.}, \textsc{YU, L.}, \textsc{Zhang, S.}, \textsc{Ghosh, G.}, \textsc{Lewis, M.}, \textsc{Zettlemoyer, L.} and \textsc{Levy, O.} (2023).
\newblock Lima: Less is more for alignment.
\newblock In \textit{Advances in Neural Information Processing Systems} (A.~Oh, T.~Naumann, A.~Globerson, K.~Saenko, M.~Hardt and S.~Levine, eds.), vol.~36. Curran Associates, Inc.

\end{thebibliography}
\bibliographystyle{ims}

\section{Appendix}

\subsection{Datasets and Expert Models}

\subsubsection{Datasets}
\label{app: dataset}
To ensure robust evaluation, we randomly sample 200 instances from each dataset as the validation set for model optimization. For the remaining data, we employ a size-based splitting strategy: for datasets with fewer than 1,000 remaining instances, we use all data as the test set; for those exceeding 1,000 instances, we retain 1,000 instances as the test set, except for multilingual tasks. For multilingual tasks (MGSM, FLORES-101), considering the scarcity of low-resource languages, we retain all remaining data as the test set to avoid potential language bias. Additionally, for FLORES-101, we create a new \textbf{FLORES-37} dataset by selecting data corresponding to the 37 languages\footnote{\href{https://cloud.google.com/vertex-ai/generative-ai/docs/learn/models}{support language}} supported by the \textit{gemma-2} model \citep{team2024gemma}. 
Furthermore, compared to the best single experiment, in the multi-task experiment, we utilize the MELD dataset, which, together with EmoryNLP, forms the tasks in the Affective Computing.

\subsubsection{Expert Models}
\begin{figure*}[htbp]
    \centering
    \includegraphics[width=\linewidth]{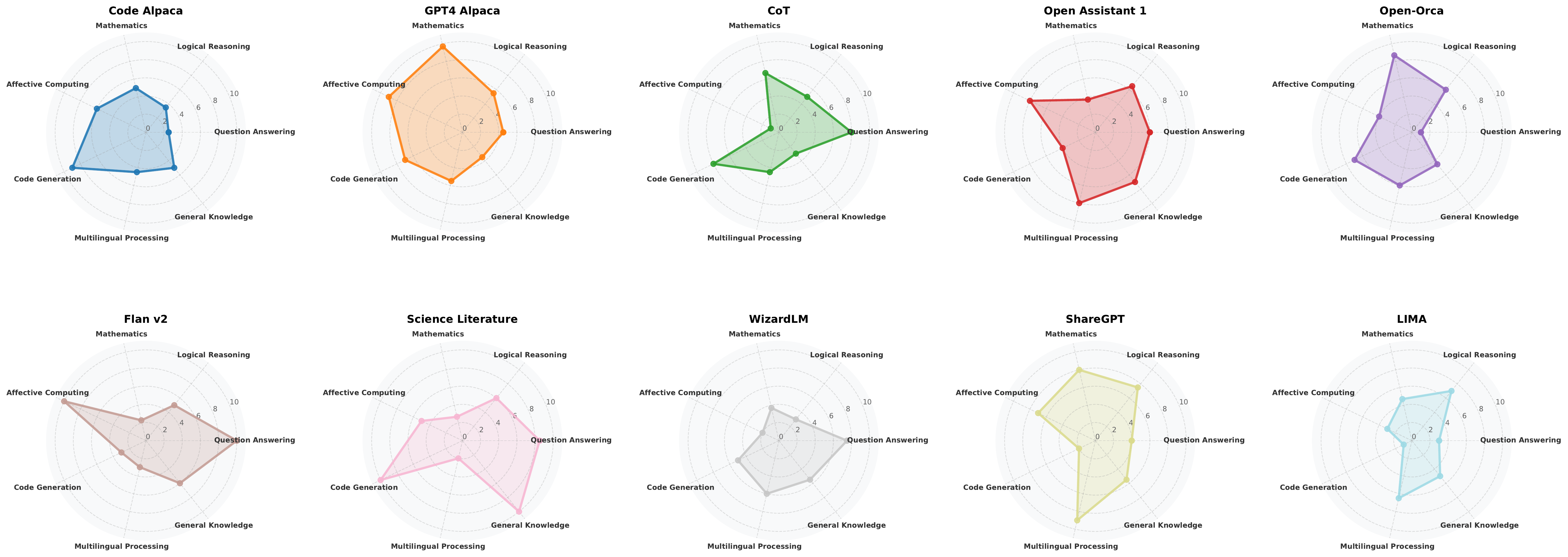}
    \caption{Capability distribution of expert models across seven dimensions, highlighting their specialized strengths.}
    \label{fig: experts}
\end{figure*}

All models are trained on 8$\times$A100-80GB GPUs.
We visualize the performances (ranks) of our ten initial expert models across all datasets, grouped by category, in Figure~\ref{fig: experts}. It is evident that each model exhibits distinct strengths, demonstrating that we have successfully trained a diverse set of experts.

\begin{table*}[htbp]
\centering
\small
\begin{tabular}{@{}lcccccc@{}}
\toprule
\textbf{Subset} & \textbf{Samples} & \textbf{LoRA rank/alpha} & \textbf{Learning rate} & \textbf{Warmup ratio} & \textbf{Batchsize} & \textbf{Epochs} \\ \midrule
CoT\citep{chung2024scaling}                 & {\color[HTML]{24292E} 49747}  & 8/16 & 2.00E-04 & 0.1 & 32 & 5 \\
Code Alpaca\citep{codealpaca}        & {\color[HTML]{24292E} 20016}  & 8/16 & 2.00E-04 & 0.1 & 32 & 5 \\
Flan v2\citep{chung2024scaling}           & {\color[HTML]{24292E} 49123}  & 8/16 & 2.00E-04 & 0.1 & 32 & 5 \\
GPT4 Alpaca\citep{peng2023instruction}        & {\color[HTML]{24292E} 19906}  & 8/16 & 2.00E-04 & 0.1 & 32 & 5 \\
Open Assistant 1\citep{kopf2024openassistant}              & {\color[HTML]{24292E} 7331}   & 8/16 & 2.00E-04 & 0.1 & 32 & 5 \\
Open-Orca\citep{mukherjee2023orca}         & {\color[HTML]{24292E} 29683}  & 8/16 & 2.00E-04 & 0.1 & 32 & 5 \\
Science Literature\citep{ivison2023camels} & {\color[HTML]{24292E} 7468}   & 8/16 & 2.00E-04 & 0.1 & 32 & 5 \\
ShareGPT\footnote{\href{https://sharegpt.com}{sharegpt}}            & {\color[HTML]{24292E} 111912} & 8/16 & 2.00E-04 & 0.1 & 32 & 1 \\
WizardLM\citep{xu2023wizardlm}            & {\color[HTML]{24292E} 29810}  & 8/16 & 2.00E-04 & 0.1 & 32 & 5 \\
LIMA\citep{NEURIPS2023_lima}                & {\color[HTML]{24292E} 1018}   & 8/16 & 2.00E-04 & 0.1 & 32 & 5 \\ \midrule
Tulu-v2-sft-mixture\citep{ivison2023tuluv2} & 326014                        & 8/16 & 2.00E-04 & 0.1 & 32 & 1 \\ \bottomrule
\end{tabular}%
\caption{Training configurations and sample sizes for different subsets. The learning rate decay follows a cosine schedule.}
\label{tab: sft parameters}
\end{table*}

\subsection{Implementation Details}
\label{sec: hyperparameters}
\subsubsection{Ensemble}
For tasks with a single correct answer, we use majority voting to aggregate predictions. For tasks without a unique answer (e.g., MBPP, Flores37/101, DROP), we apply a similarity-based approach. Specifically, we take the top-k outputs, embed each using \texttt{text-embedding-3-small}, and compute pairwise similarities via BERTScore~\citep{zhang2019bertscore}. The output with the highest overall similarity score is selected as the final ensemble result.

\subsubsection{Baseline}
\label{sec: baseline detailed}
To ensure fair comparisons, we use the same set of 10 initial expert models for all baselines except for the data merge baseline, and we otherwise align our settings as closely as possible with the original papers. LoRAHub is run for 50 iterations, and Model Swarms uses a population size of 10 for 10 iterations, with other hyperparameters set according to its original recommendations. Unless otherwise specified, all experiments are performed on 4$\times$A100 80G GPUs and repeated 5 times.

\subsubsection{Prompts}
\begin{figure}[htbp]
    \centering
    \begin{minipage}[b]{0.48\textwidth}
         \centering
         \includegraphics[width=\linewidth]{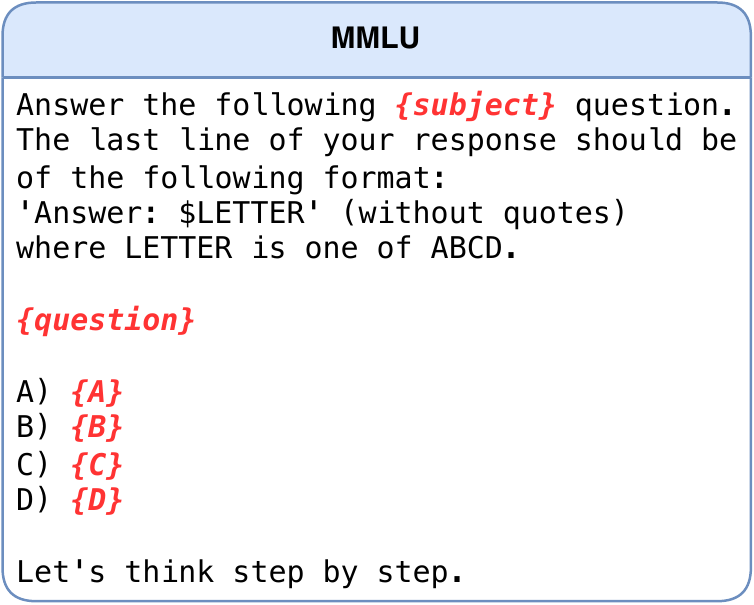}
         \caption{The prompt of MMLU.}
         \label{fig:mmlu}
    \end{minipage}
    \hfill
    \begin{minipage}[b]{0.48\textwidth}
         \centering
         \includegraphics[width=\linewidth]{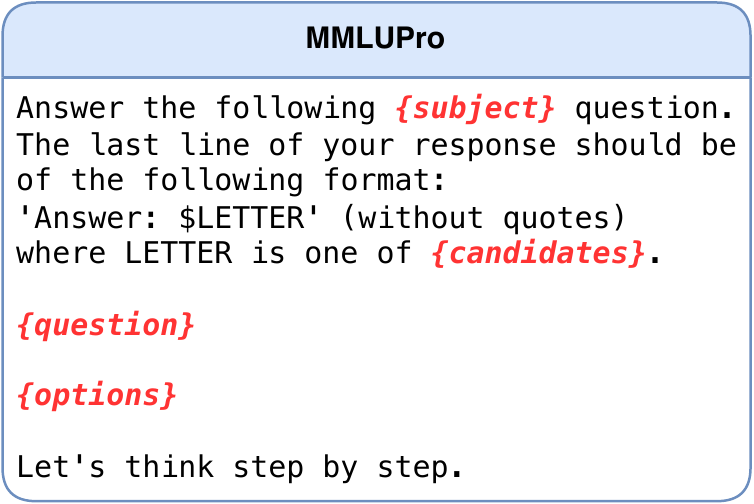}
         \caption{The prompt of MMLUPro.}
         \label{fig:mmlupro}
    \end{minipage}
\end{figure}

\begin{figure}[htbp]
    \centering
    \begin{minipage}[b]{0.48\textwidth}
        \centering
        \includegraphics[width=\linewidth]{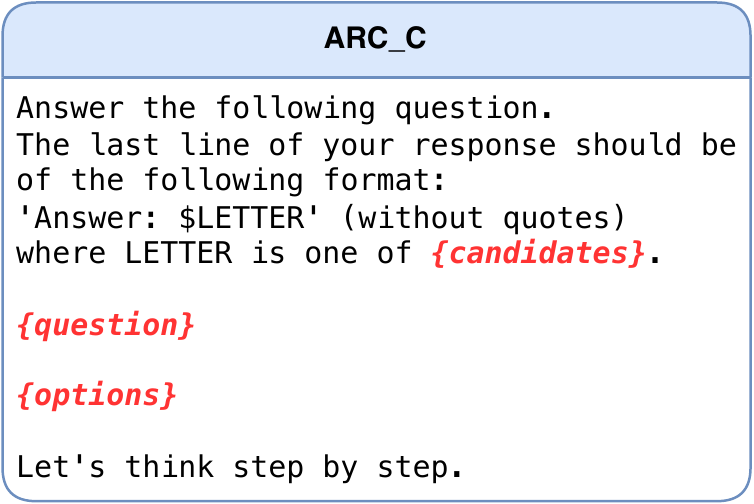}
        \caption{The prompt of ARC\_C.}
        \label{fig: arc_c prompt}
    \end{minipage}
    \hfill
    \begin{minipage}[b]{0.48\textwidth}
        \centering
        \includegraphics[width=\linewidth]{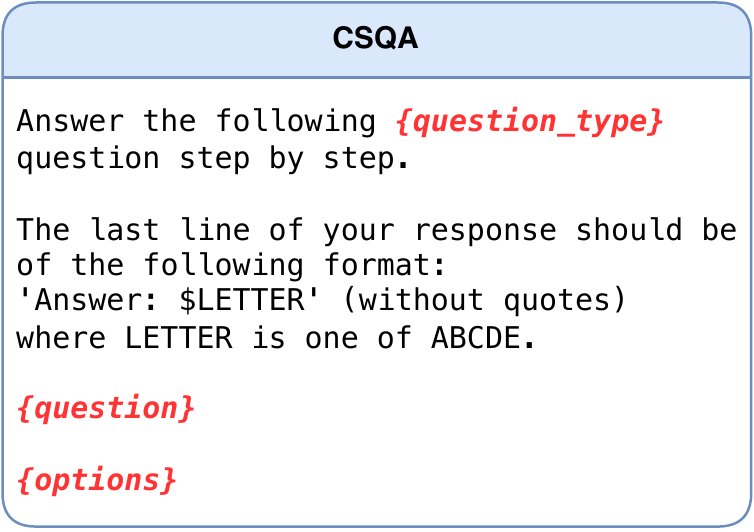}
        \caption{The prompt of CSQA.}
        \label{fig: csqa prompt}
    \end{minipage}
\end{figure}

\begin{figure}[htbp]
    \centering
    \begin{minipage}[b]{0.48\textwidth}
        \centering
        \includegraphics[width=\linewidth]{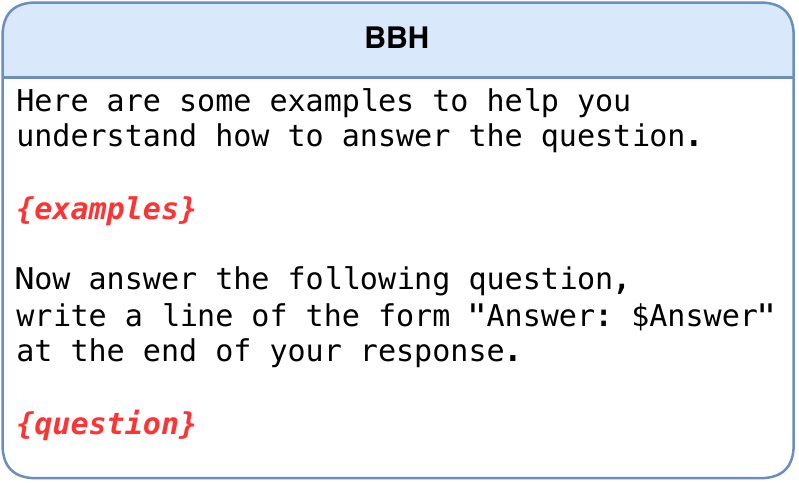}
        \caption{The prompt of BBH.}
        \label{fig: bbh prompt}
    \end{minipage}
    \hfill
    \begin{minipage}[b]{0.48\textwidth}
        \centering
        \includegraphics[width=\linewidth]{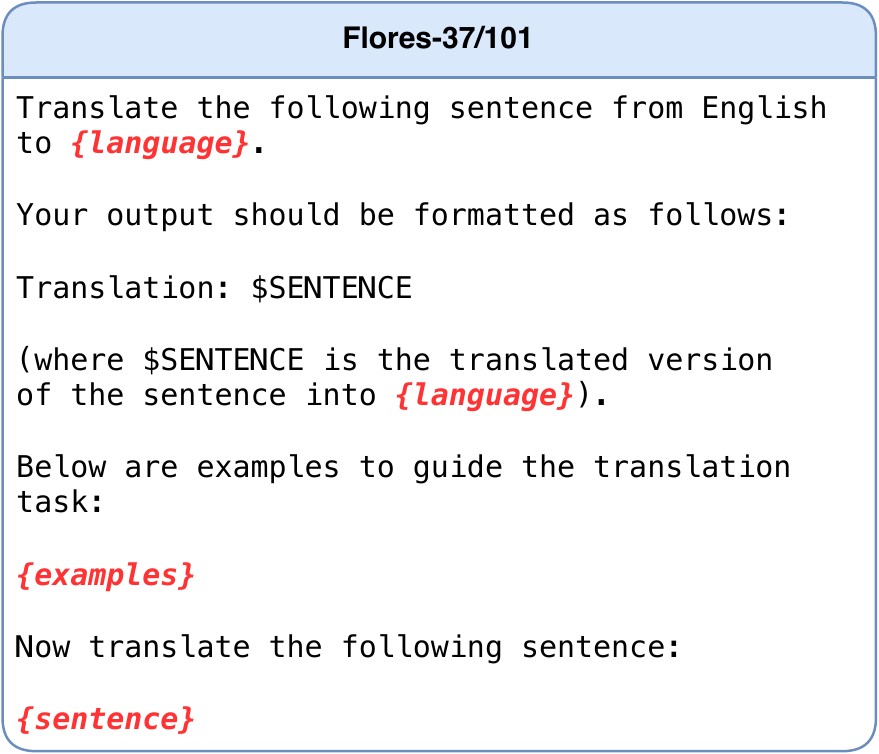}
        \caption{The prompt of Flores-37/101.}
        \label{fig: flores prompt}
    \end{minipage}
\end{figure}

\begin{figure}[htbp]
\centering
    \begin{minipage}[b]{0.48\textwidth}
        \centering
        \includegraphics[width=\linewidth]{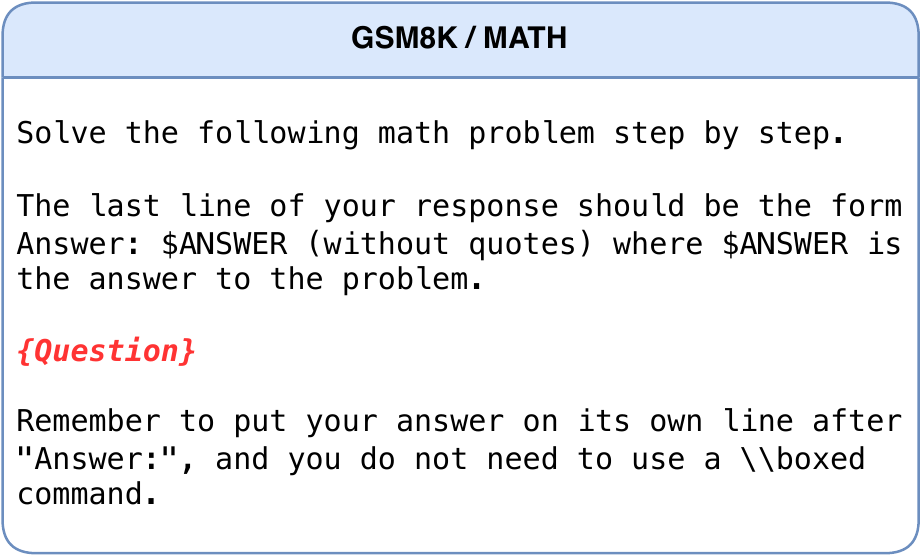}
        \caption{The prompt of GSM8K and MATH.}
        \label{fig: gsm8k/math prompt}
    \end{minipage}
    \hfill
    \begin{minipage}[b]{0.48\textwidth}
        \centering
        \includegraphics[width=\linewidth]{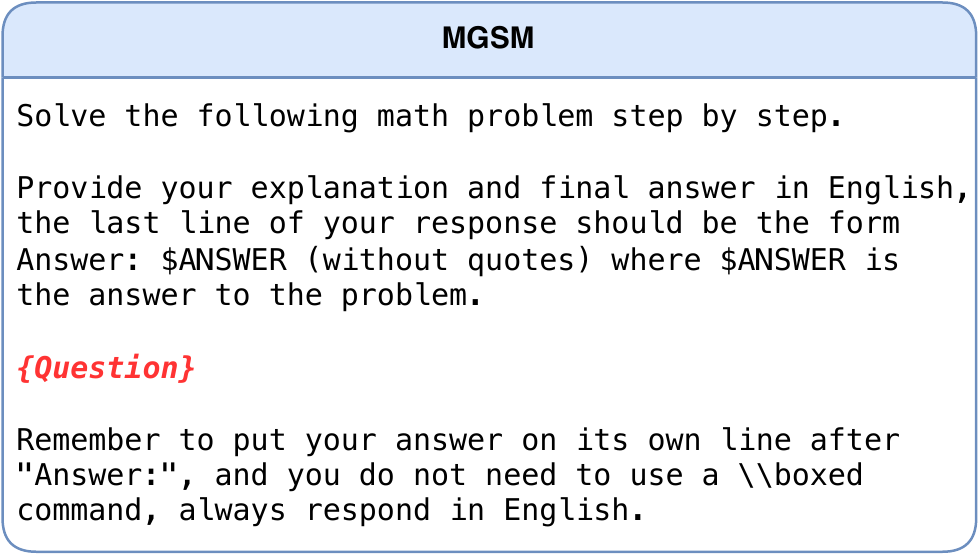}
        \caption{The prompt of MGSM.}
        \label{fig: mgsm prompt}
    \end{minipage}
\end{figure}

\begin{figure}
    \centering
    \begin{minipage}[b]{0.48\textwidth}
        \centering
        \includegraphics[width=\linewidth]{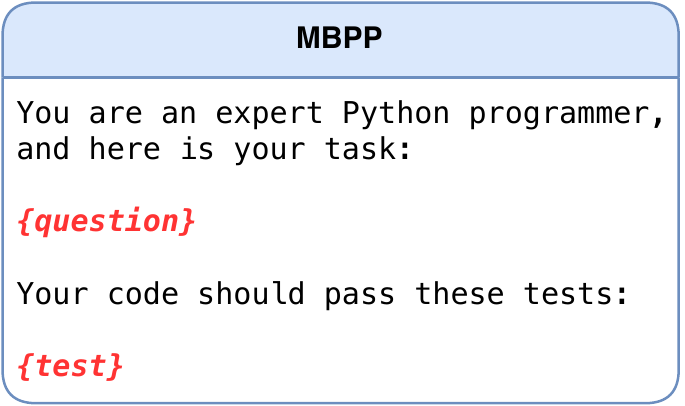}
        \caption{The prompt of MBPP.}
        \label{fig: mbpp prompt}
    \end{minipage}
    \hfill
    \begin{minipage}[b]{0.48\textwidth}
        \centering
        \includegraphics[width=\linewidth]{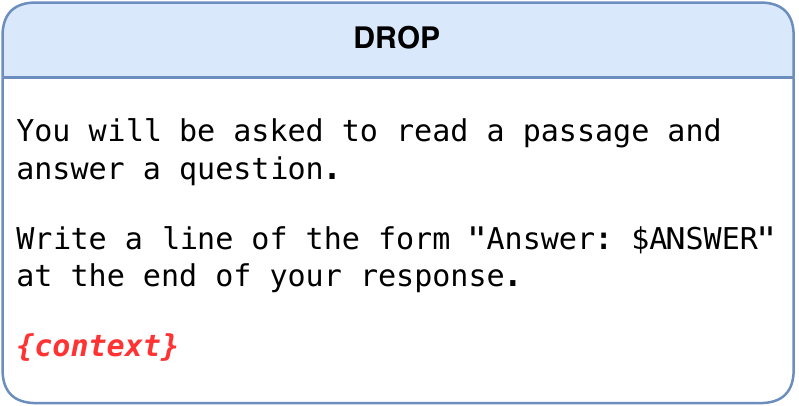}
        \caption{The prompt of DROP.}
        \label{fig: drop prompt}
    \end{minipage}
\end{figure}

\begin{figure}[htbp]
    \centering
    \includegraphics[width=0.8\linewidth]{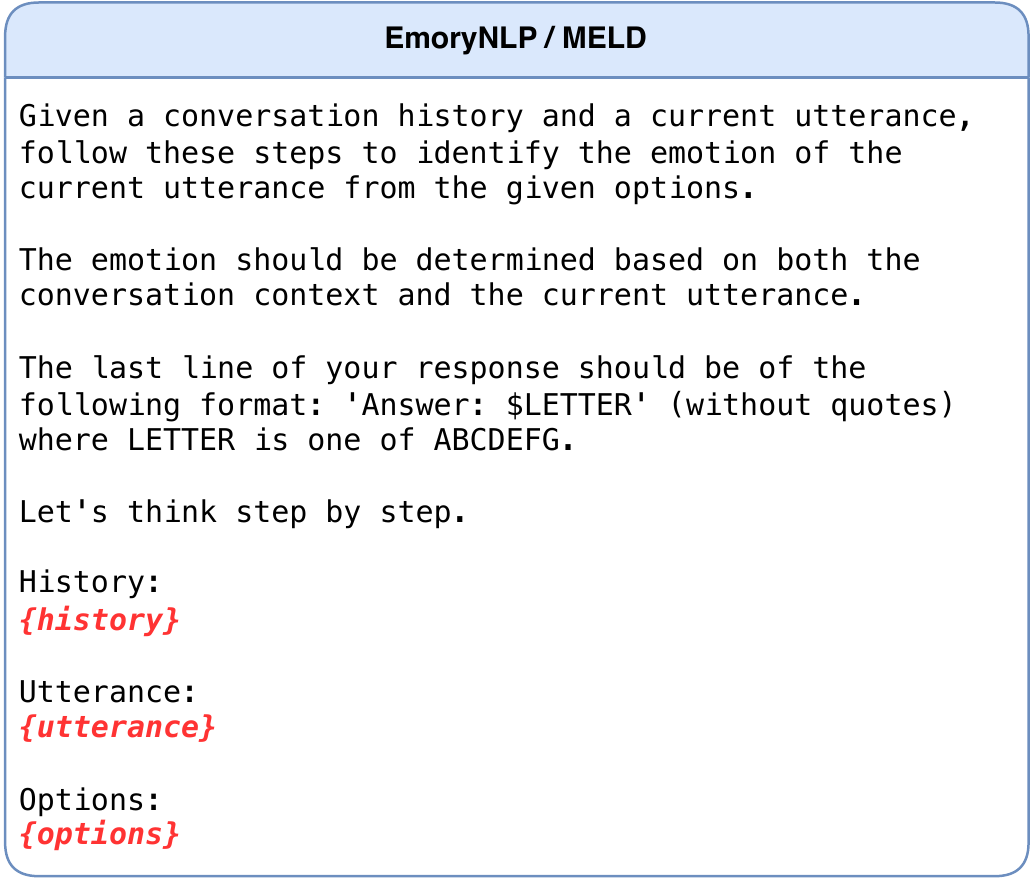}
    \caption{The prompt of EmoryNLP and MELD.}
    \label{fig: emorynlp and meld prompt}
\end{figure}

All methods in our experiments rely on the same prompts for a fair comparison.
Following OpenAI's approach\footnote{\url{https://github.com/openai/simple-evals}}, we design specific prompts for different tasks while maintaining a minimalist design philosophy. 
All prompts are shown in Figure \ref{fig:mmlu}-\ref{fig: emorynlp and meld prompt}.
As MMLUPro questions can have more than four options, we implement a more flexible prompt structure (Figure \ref{fig:mmlupro}) to accommodate the varying number of choices.

\subsection{Hyperparameter Statistical Analysis}
\label{sec: hyper analysis}
To optimize hyperparameter selection and verify the stability of the \ga and \gsa methods, we conduct 500 experiments. The hyperparameters involved in the study include:
\begin{itemize}
    \item Cross rate ($c_r$): Range from 0.1 to 1.0.
    \item Individual mutation rate ($im_r$): Range from 0.1 to 1.0.
    \item Gene mutation rate ($gm_r$): Range from 0.1 to 1.0.
\end{itemize}

The dependent variables are \textit{test performance}, \textit{Top-3 ensemble performance}, and \textit{optimization time}. The experiments use a fixed random seed of 47, set the population size (N) to 10, $\sigma$ to 0.001, and selected \ga as the test method (we believe that the hyperparameter settings of \ga and \gsa are similar). The dataset use in this experiment is MMLUpro, which is a highly comprehensive test dataset. The experimental environment consisted of $4\times A100$ 80G GPUs. In each experiment, we discretize each hyperparameter range into fixed intervals of 0.1 (e.g., 0.1, 0.2, 0.3, $\dots$, 1.0). From these discrete candidate values, one value was randomly selected for $c_r$, $im_r$, and $gm_r$ in each experimental run, subject to the same random seed initialization. 

In the correlation analysis (see in Table \ref{tab: correlation}), Spearman and Pearson methods reveal that the $c_r$ had a significant positive correlation with \textit{test performance} and \textit{optimization time}, and a weak correlation with \textit{ensemble performance}. The $im_r$ is found to be weakly negatively correlated with \textit{test performance} but significantly positively correlated with \textit{optimization time}. The $gm_r$ show small correlation coefficients with all dependent variables. 

\begin{table}[htbp]
\centering
\small
\begin{tabular}{llcccc}
\toprule
\multirow{2.5}{*}{\textbf{HyperParameter}} & \multirow{2}{*}{\textbf{Metric}} & \multicolumn{2}{c}{\textbf{Pearson}} & \multicolumn{2}{c}{\textbf{Spearman}} \\
\cmidrule(lr){3-4} \cmidrule(lr){5-6}
& & \textbf{$r$} & \textbf{p-value} & \textbf{$\rho$} & \textbf{p-value} \\
\midrule
\multirow{3}{*}{Cross Rate} 
& Test & 0.192*** & 2.132e-03 & 0.194*** & 1.927e-03 \\
& Ensemble & 0.119* & 5.831e-02 & 0.102 & 1.052e-01 \\
& Time & 0.382*** & 3.030e-10 & 0.408*** & 1.395e-11 \\
\midrule
\multirow{3}{*}{Individual Mutation Rate} 
& Test & -0.134** & 3.284e-02 & -0.118* & 5.946e-02 \\
& Ensemble & -0.091 & 1.479e-01 & -0.094 & 1.368e-01 \\
& Time & 0.775*** & 4.647e-52 & 0.796*** & 9.183e-57 \\
\midrule
\multirow{3}{*}{Gene Mutation Rate}
& Test & -0.093 & 1.395e-01 & -0.092 & 1.433e-01 \\
& Ensemble & 0.077 & 2.240e-01 & 0.092 & 1.423e-01 \\
& Time & 0.123* & 5.002e-02 & 0.125** & 4.684e-02 \\
\bottomrule
\multicolumn{6}{l}{\textit{Notes:} *** $p<0.01$, ** $p<0.05$, * $p<0.1$} \\
\end{tabular}
\caption{Correlation Analysis between Algorithm Parameters and Performance Metrics}
\label{tab: correlation}
\end{table}

Given the complex coupling relationships between variables (e.g., gene mutation rate is only meaningful when individual mutation rate is greater than 0), simple correlation analysis may not be sufficient. Therefore, we conduct further multiple linear regression analysis. For example, for \textit{test performance}, we construct the following regression model:
\begin{equation}
    \text{test}_i = \beta_0 + \beta_1 \cdot c_{ri} + \beta_2 \cdot im_{ri} + \beta_3 \cdot gm_{ri} + \epsilon_i
\end{equation}

We construct corresponding regression models for \textit{ensemble performance} and \textit{optimization time} and estimate the parameters using the least squares method. The significance test results are shown in Table \ref{tab: regression_results}.

Analysis results show that $c_r$ has a significant positive impact on \textit{test performance}, and $im_r$ shows a negative impact. The impact of $gm_r$ do not reach significance. For \textit{ensemble performance}, both correlation and multiple regression analyses show that the effects of the three hyperparameters are not significant, indicating that after adopting the ensemble strategy, the \ga's sensitivity to hyperparameters decreased. In terms of \textit{optimization time}, increased cross rate and individual mutation rate significantly prolonged total runtime.

\begin{table}[htbp]
\centering
\small
\begin{tabular}{lrrrr}
\toprule
\textbf{Variable} & \textbf{Coefficient} & \textbf{Std. Error} & \textbf{t-statistic} & \textbf{p-value} \\
\midrule
\multicolumn{5}{c}{\textbf{Panel A: Test Performance}} \\
\midrule
Constant & 0.2881*** & 0.001 & 198.057 & 0.000 \\
Cross Rate & 0.0054*** & 0.002 & 3.279 & 0.001 \\
Individual Mutation Rate & -0.0030** & 0.002 & -1.974 & 0.049 \\
Gene Mutation Rate & -0.0028* & 0.002 & -1.826 & 0.069 \\
\midrule
\multicolumn{5}{c}{\textbf{Panel B: Ensemble Performance}} \\
\midrule
Constant & 0.2985*** & 0.002 & 150.732 & 0.000 \\
Cross Rate & 0.0038* & 0.002 & 1.700 & 0.090 \\
Individual Mutation Rate & -0.0030 & 0.002 & -1.455 & 0.147 \\
Gene Mutation Rate & 0.002 & 0.002 & 1.071 & 0.285 \\
\midrule
\multicolumn{5}{c}{\textbf{Panel C: Computational Time (s)}} \\
\midrule
Constant & 507.28*** & 116.53 & 4.353 & 0.000 \\
Cross Rate & 1724.37*** & 130.94 & 13.169 & 0.000 \\
Individual Mutation Rate & 3159.24*** & 122.64 & 25.760 & 0.000 \\
Gene Mutation Rate & 111.91 & 122.96 & 0.910 & 0.364 \\
\bottomrule
\multicolumn{5}{l}{\textit{Notes:} *** $p<0.01$, ** $p<0.05$, * $p<0.1$} \\
\end{tabular}
\caption{Regression Results for Genetic Algorithm Parameters}
\label{tab: regression_results}
\end{table}

Combining the results of correlation and regression analysis, it can be concluded that the $c_r$ plays a significant role in improving the quality of solutions (test performance) generated by \ga, though it also increases computational runtime. This finding suggests that a higher $c_r$ enhances the combination quality of chromosomes (model weights), leading to better solutions, albeit at the cost of greater computational effort. On the other hand, an excessively high $im_r$ tends to disrupt superior individuals and substantially enlarge the search space, leading to increased time consumption. However, a moderate level of $im_r$ proves beneficial as it helps prevent the population from prematurely converging. In contrast, the $gm_r$ demonstrate only a limited impact within the scope of this study, implying its influence on the \ga's performance is relatively minor.

Within a broad range of hyperparameter settings, the performance of \ga is relatively stable, suggesting it is not sensitive to hyperparameter values. The introduction of the ensemble strategy further reduces sensitivity, effectively balancing different parameter choices while maintaining solution quality.

The recommended hyperparameter settings are a $c_r$ of 0.3 to 0.6, $im_r$ of 0.1, and $gm_r$ of 0.1. This setup can build superior individuals while maintaining population diversity and avoiding the negative effects of excessive mutation. Even in larger-scale or more complex problem scenarios, this configuration, through the ensemble strategy, can smooth out parameter differences and fully leverage the core advantages of \ga, achieving relatively stable overall performance.

\end{document}